\def\preprintmode{}  %
\documentclass[11pt]{article}
\ifdefined\preprintmode
\usepackage[preprint]{acl}
\else
\usepackage[review]{acl}
\fi
\usepackage{times}
\usepackage{latexsym}
\usepackage[T1]{fontenc}
\usepackage[utf8]{inputenc}
\usepackage{microtype}
\IfFileExists{upquote.sty}{\usepackage{inconsolata}}{}  %
\usepackage{amsmath,amssymb}
\usepackage{graphicx}
\usepackage{subcaption}
\usepackage{booktabs}
\usepackage{longtable}
\usepackage{array}
\usepackage[table]{xcolor}  %
\usepackage{enumitem}
\usepackage{fancyvrb}  %
\newsavebox{\jsonbox}
\usepackage{graphicx}   %
\usepackage{subcaption} %
\usepackage{wrapfig}    %
\usepackage{fontawesome}  %

\usepackage{adjustbox}

\newcommand{\nDomains}{16}
\newcommand{\nEnvFamilies}{18}
\newcommand{\nEnvSuites}{22}
\newcommand{\nEnvs}{138}

\newcommand{\nLeaves}{105}

\newcommand{\nLoaderOnly}{2,607}
\newcommand{\nQuestions}{92,125}

\newcommand{\nSoft}{4,425}
\newcommand{\nSources}{268}

\newcommand{\pctTierC}{33}

\newcommand{\nQuestionsThree}{78,362}
\newcommand{\nLeavesThree}{87}

\definecolor{topone}{HTML}{8FB8EC}
\definecolor{toptwo}{HTML}{BCD5F4}
\definecolor{topthree}{HTML}{E2ECFA}
\definecolor{zeroone}{HTML}{8CD3C0}
\definecolor{zerotwo}{HTML}{BFE7DC}
\definecolor{zerothree}{HTML}{E0F3EE}
\definecolor{overallcol}{HTML}{EFEFEF}  %
\definecolor{better}{HTML}{009E73}
\definecolor{worse}{HTML}{D55E00}

\newcommand{\name}{System1-Arena}

\newcommand{\codeurl}{https://github.com/TianyuCodings/S1-Arena}
\newcommand{\siteurl}{https://tianyucodings.github.io/S1-Arena}
\newcommand{\dataurl}{https://huggingface.co/datasets/C-Tianyu/S1-Arena}
\newcommand{\hficon}{\raisebox{-0.15em}{\includegraphics[height=1.0em]{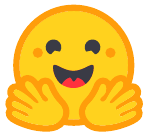}}}

\title{\name: Which Decisions Can Be Left to System 1?}
\ifdefined\preprintmode
\author{Tianyu Chen\thanks{Equal contribution.} \quad Xinyang Liu\footnotemark[1] \quad Ruyi An \quad Zhendong Wang \quad
  Mingyuan Zhou\thanks{\raggedright Corresponding author: \texttt{mingyuan.zhou@mccombs.utexas.edu}} \\
  The University of Texas at Austin}
\else
\author{Tianyu Chen \quad Xinyang Liu \quad Ruyi An \quad Zhendong Wang \quad Mingyuan Zhou \\
  The University of Texas at Austin}  %
\fi

\begin{document}
\maketitle

\begin{abstract}
Large language models (LLMs) often make discrete decisions, such as flagging phishing or escalating to a human. \emph{System-1 models}, such as Jev and the OpenAI Decisions API, return probabilities over options in a single forward pass, while \emph{System-2} approaches generate reasoning steps. Which decisions can System-1 models handle, and which require generative reasoning? We introduce \name{}, a benchmark that gives all models the same decision requests and scores their probabilities. It contains 92k questions across \nLeaves{} tasks from \nSources{} public sources, in four groups: text classification and knowledge, applied decisions, evaluation and control, and agent and action. It also includes \nEnvs{} closed-loop environments, from web shopping and SWE-bench coding control to robotics, games, and driving; in real-time mode, the world evolves while a model decides. We evaluate 30 models: 7 closed LLMs, 13 open LLMs read at the answer token, and 10 System-1 models. The largest System-1 models match or exceed fast closed LLMs in non-agentic accuracy, with better-calibrated probabilities at one-quarter of the latency. A reasoning LLM is the most accurate, at about $70\times$ the cost of a fast LLM. In closed loops, large open LLMs succeed most often (up to 32.9\%); small System-1 models rarely succeed.
\end{abstract}

\ifdefined\preprintmode
\begin{center}
\small
\href{\codeurl}{\faGithub~Code} \quad
\href{\dataurl}{\hficon~Dataset} \quad
\href{\siteurl}{\faGlobe~Leaderboard}
\end{center}
\fi
\section{Introduction}
\label{sec:intro}

\begin{figure}[t]
\centering
\includegraphics[width=\columnwidth]{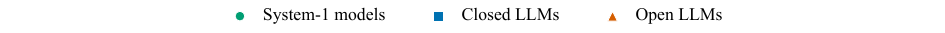}\par
\captionsetup[subfigure]{
  font=footnotesize,
  labelfont=bf,
  justification=centering,
  singlelinecheck=false,
  skip=3pt
}
\begin{subfigure}[t]{0.48\columnwidth}
  \centering
  \includegraphics[width=\linewidth]{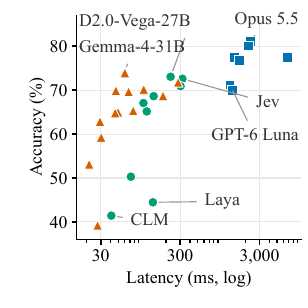}
  \caption{Fast decisions\\can be accurate.}
  \label{fig:front-accuracy}
\end{subfigure}\hfill
\begin{subfigure}[t]{0.48\columnwidth}
  \centering
  \includegraphics[width=\linewidth]{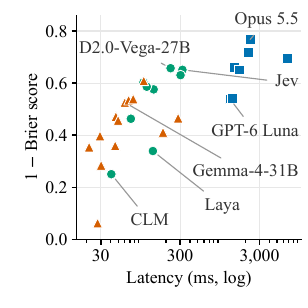}
  \caption{Probability quality\\changes model rankings.}
  \label{fig:front-probability}
\end{subfigure}
\caption{\textbf{System-1 models combine fast, accurate decisions with well-calibrated probabilities.}
The strongest System-1 models match fast closed LLMs in accuracy, with better probability estimates and lower latency.
Reasoning LLMs achieve the highest accuracy.
See Section~\ref{sec:setup} for evaluation settings.}
\label{fig:front}
\end{figure}

Many applications ask large language models (LLMs) to make discrete decisions.
Examples include detecting phishing, deciding whether to involve a human, checking task completion, and selecting the next action.
Such requests often occur at high volume and need a label or option with a probability estimate.
Reasoning LLMs can spend unnecessary computation on simple decisions \citep{zhang2025s1bench}.
Routing and cascading reduce these costs, but they still rely on generative models \citep{chen2023frugalgpt,hu2024routerbench}.
Latency also limits LLM use in robotics, driving, and games, where the world continues to change during inference.

\emph{System~1} supports fast, automatic judgments, while \emph{System~2} supports slower, deliberate reasoning \citep{kahneman2011thinking}.
LLM studies often use this distinction to describe shorter and longer reasoning by the same model \citep{yu2024distilling,ziabari2025spectrum,zhang2025s1bench}.
Here, we study models that directly predict probabilities for discrete decisions.
Jev \citep{typesafe2026jev} and the Decision-2.0 family \citep{vllmsr2026decision2} accept a state and typed questions.
Question types include yes/no, multiple choice, and ordered scales.
These models return a probability distribution for each question in one forward pass, without generating text. The OpenAI Decisions API \citep{openai2026decisions} provides the same decision interface through GPT-6 Luna, and Microsoft-Decision-1 \citep{microsoft2026decision1} provides it through a post-trained Qwen3.5-9B.
These systems can respond within a few hundred milliseconds at low cost.
In our runs, Jev cost about \$0.05 per 1,000 requests.
Probability estimates support decision thresholds, calibration, and rules for escalation to a human.

This raises a central question: which decisions can System-1 models handle, and where do generative LLMs perform better?
Existing benchmarks provide limited evidence.
JevBench \citep{benchmarkheaven2026jevbench} and ReflexBench \citep{brida2026reflexbench} each contain only a few hundred synthetic decisions.
Their narrow coverage makes model rankings sensitive to task selection.
They mainly compare System-1 models and do not include agentic tasks.
These tasks are an important use case because each decision can affect later actions, and delays can change outcomes.

We introduce \name{}, a benchmark with \nQuestions{} typed questions across \nLeaves{} tasks from \nSources{} public sources in \nDomains{} domains.
It also includes \nEnvs{} closed-loop environments from \nEnvFamilies{} families.
In real-time mode, the environment continues to evolve while a model decides.
We evaluate System-1 models alongside open and closed LLMs through a shared decision interface.
Each model family uses a fixed prompt, and we score the predicted probabilities. \name{} follows three design choices.
\begin{itemize}[leftmargin=*,itemsep=0pt,topsep=0pt]
  \item \textbf{Broad coverage.}
  Every model receives the same decision request for each example: a state and typed questions.
  Labels come from test execution, solvers, experts, annotator distributions, and resolved events.
  We preserve the instructions from each source (Section~\ref{sec:static}).
  \item \textbf{More than accuracy.}
  Each model returns a probability distribution for every question.
  We report accuracy and use the Brier score to assess probability quality.
  More details are in Section~\ref{sec:metrics}.
  \item \textbf{Agentic tasks.}
  We evaluate individual agent decisions and success in closed-loop environments.
  In real-time mode, the environment continues to evolve for the duration of each model call, using that model's measured latency (Section~\ref{sec:loop}).
\end{itemize}

We evaluate 30 models: 7 closed LLMs, 13 open LLMs evaluated through answer-token probabilities, and 10 System-1 models.
On non-agentic tasks, Jev and Decision-2.0-Vega-27B reach 72.6\% and 73.0\% accuracy, respectively, compared with 71.2\% for GPT-6 Luna.
They also produce better-calibrated probabilities at lower latency than fast closed LLMs (Figure~\ref{fig:front}).
The performance gap between the best System-1 models and the best LLMs is smallest on filtering, grouping, ranking, retrieval, and dialogue tasks. System-1 models remain less accurate on knowledge and forecasting tasks (Figure~\ref{fig:front-agentic}).

Reasoning LLMs achieve the highest accuracy.
Claude Opus 5.5 reaches 81.1\%, at about $70\times$ the cost of a fast LLM.
Explicit reasoning also improves open LLM accuracy.
With a few hundred reasoning tokens per question, Qwen3.8-27B reaches 75.6\%, exceeding Jev's accuracy but producing less-calibrated probabilities (Figure~\ref{fig:think}).

In agentic tasks, some models perform well on single-step decision tasks but struggle with long-horizon tasks. Decision-2.0-Vega-27B, Jev, and Qwen3.5-9B all achieve 67.8\% accuracy on single-step tasks. On long-horizon tasks, Gemma-4-31B achieves the highest success rate at 32.9\%. The four small System-1 models reach at most 3.1\%, showing limited ability to handle long-horizon planning. The OpenAI Decisions API is faster and better calibrated than GPT-6 Luna, but achieves a lower success rate on long-horizon tasks.

In sum, we make four contributions.
First, \name{} provides a shared decision interface for System-1 models and LLMs, with labels and instructions linked to their sources.
Second, our real-time protocol advances the environment during model inference to measure the value of fast decisions.
Third, we provide a method for obtaining decision probabilities from open autoregressive and diffusion LLMs through their answer tokens.
Fourth, our evaluation of 30 models identifies where System-1 models match LLMs and where they fall behind.

\section{Related Work}
\label{sec:related}

\paragraph{System~1 and System~2 in language models.}
The dual-process distinction of \citet{kahneman2011thinking} is often used to describe how much a language model reasons before answering.
\citet{yu2024distilling} distill long reasoning outputs into models that answer directly.
\citet{ziabari2025spectrum} align models toward either end of this spectrum and find that faster responses perform better on commonsense tasks, while slower reasoning performs better on arithmetic and symbolic tasks.
S1-Bench \citep{zhang2025s1bench} shows that large reasoning models overthink simple questions.
These studies examine System~1 and System~2 behavior within generative models.
\name{} compares generative LLMs with a separate class of models that return decision probabilities without generating text, and identifies which decisions these models can handle.

\paragraph{Typed decision models and zero-shot classifiers.}
Zero-shot text classification assigns labels specified at inference time.
\citet{yin2019zeroshot} frame this task as entailment, and \citet{laurer2023universal} train universal classifiers across tasks formatted as natural language inference.
System-1 models such as Jev \citep{typesafe2026jev}, Microsoft-Decision-1 \citep{microsoft2026decision1}, and the OpenAI Decisions API extend this interface to multiple typed questions per input---boolean, choice, and ordered scores---with a probability for each option.
JevBench \citep{benchmarkheaven2026jevbench} and ReflexBench \citep{brida2026reflexbench} evaluate accuracy, calibration, robustness, and speed on a few hundred newly written decisions, mainly comparing System-1 models.
\name{} covers 92k questions with source-provided labels, compares System-1 models and LLMs on the same decision requests, and evaluates them in closed-loop environments.

\paragraph{Holistic benchmarks, calibration, and cost.}
MMLU \citep{hendrycks2021mmlu}, BIG-bench \citep{srivastava2022bigbench}, and HELM \citep{liang2022helm} evaluate models across diverse tasks using a common protocol.
HELM also reports calibration and efficiency.
The Brier score \citep{brier1950verification} evaluates probabilistic predictions, while expected calibration error \citep{guo2017calibration} measures the gap between confidence and accuracy.
Large models can produce well-calibrated token probabilities on multiple-choice questions \citep{kadavath2022know}.
For instruction-tuned models, verbalized probabilities can be better calibrated than token probabilities \citep{tian2023calibration}, but can still be overconfident \citep{xiong2024uncertainty}.
We use verbalized probabilities for closed LLMs to obtain a full distribution for each question through their text-generation APIs.
Cost-aware methods route queries among LLMs of different sizes \citep{chen2023frugalgpt,hu2024routerbench}.
\name{} asks which decisions a $\text{System-1}$ model can handle, informing routing between System-1 models and generative LLMs.

\paragraph{Interactive and latency-sensitive evaluation.}
Most benchmarks for web agents, text games, customer support, coding, and reinforcement learning pause the environment while the model decides.
Appendix~\ref{app:envs} lists the environments used in our evaluation.
\citet{kang2025winfast} show that tailoring model size and quantization to a latency budget improves performance in both a trading simulation and a fighting game.
We use a harness that presents each environment's decisions as typed questions and add a real-time mode that incorporates each model's measured latency into the environment dynamics.

\section{\name{}}
\label{sec:benchmark}

\subsection{Decision requests}
\label{sec:interface}

We represent every static example and environment step as a \emph{decision request} $r = (s, Q)$ (Figure~\ref{fig:request}).
The state $s$ contains a small set of named text fields.
Examples include a message, a code diff with its commit message, or a rendered web page.
The questions $Q = \{q_1, \dots, q_m\}$ share this state.

Each question is \textbf{boolean} (yes/no), \textbf{choice} (2--255 options with exactly one correct answer), or \textbf{score} (2--10 ordered levels).
Score levels can represent a grading rubric.
Each question includes instructions, and each option includes a description, so models need no dataset-specific prompt.
A model returns a probability distribution over the options for each question.
The reference answer is a hard label or, when available, a distribution provided by the source.

\definecolor{jsonkey}{HTML}{1D5FA8}
\definecolor{jsonstr}{HTML}{0F766E}
\definecolor{jsonnum}{HTML}{C2410C}
\definecolor{jsoncom}{HTML}{6B7280}
\definecolor{jsonbg}{HTML}{F5F7FA}
\definecolor{jsonrule}{HTML}{D0D7E2}
\definecolor{jsongold}{HTML}{FFE58A}
\providecommand{\jk}[1]{\textcolor{jsonkey}{#1}}
\providecommand{\js}[1]{\textcolor{jsonstr}{#1}}
\providecommand{\jn}[1]{\textcolor{jsonnum}{#1}}
\providecommand{\jc}[1]{\textcolor{jsoncom}{\itshape #1}}
\providecommand{\jg}[1]{{\setlength{\fboxsep}{0.7pt}\colorbox{jsongold}{#1}}}
\providecommand{\LB}{\char`\{}
\providecommand{\RB}{\char`\}}
\providecommand{\jsonhead}[2]{\noindent{\small\textbf{#1}}\hfill{\scriptsize\textcolor{jsoncom}{#2}}\par\vspace{2pt}}
\providecommand{\jsonframe}{\setlength{\fboxsep}{3pt}\fcolorbox{jsonrule}{jsonbg}{\usebox{\jsonbox}}}

\begin{figure}[t]
\jsonhead{Decision request}{a state and typed questions}
\begin{lrbox}{\jsonbox}\begin{minipage}{\dimexpr\linewidth-6.8pt\relax}
\begin{Verbatim}[commandchars=\\\{\},fontsize=\fontsize{7}{8.1}\selectfont]
\LB
  \jk{"state"}: \LB
    \jk{"vulnerability_description"}: \js{"SQL Injection}
\js{      vulnerability in Limesurvey v2.65.1+170522 ...}
\js{      via 'token' parameter in '/index.php' endpoint."}
  \RB,
  \jk{"questions"}: \LB
    \jk{"attack_vector"}: \LB
      \jk{"type"}: \js{"choice"},
      \jk{"instructions"}: \js{"Attack Vector (AV): ..."},
      \jk{"criteria"}: \LB\jk{"local"}: \js{"..."}, \jk{"physical"}: \js{"..."},
                   \jk{"network"}: \js{"..."}, \jk{"adjacent"}: \js{"..."}\RB
    \RB,
    \jk{"severity"}: \LB
      \jk{"type"}: \js{"score"},
      \jk{"instructions"}: \js{"Which CVSS v3.1 severity ...?"},
      \jk{"criteria"}: [\js{"None: ..."}, \js{"Low: ..."}, \js{"Medium: ..."},
                   \js{"High: ..."}, \js{"Critical: ..."}]
    \RB,
    ...                              \jc{// 7 more questions}
  \RB
\RB
\end{Verbatim}
\end{minipage}\end{lrbox}%
\jsonframe

\vspace{5pt}
\jsonhead{Answer of Jev}{a distribution per question; gold option marked}
\begin{lrbox}{\jsonbox}\begin{minipage}{\dimexpr\linewidth-6.8pt\relax}
\begin{Verbatim}[commandchars=\\\{\},fontsize=\fontsize{7}{8.1}\selectfont]
\LB
  \jk{"attack_vector"}: \LB\jg{\jk{"network"}}: \jn{1.00}, \jk{"local"}: \jn{0.00}, ...\RB,
  \jk{"severity"}: \LB\jg{\jk{"4"}}: \jn{0.76}, \jk{"3"}: \jn{0.19}, \jk{"2"}: \jn{0.02}, ...\RB,
  ...                                \jc{// "4" = Critical}
\RB
\end{Verbatim}
\end{minipage}\end{lrbox}%
\jsonframe
\caption{\textbf{A decision request from the CVSS task and the answer of Jev.} We show two of its nine questions and
shorten the instructions and the option descriptions. An LLM gets the same request as text (Appendix~\ref{app:prompt}).}
\label{fig:request}
\vspace{-0.5cm}
\end{figure}

System-1 models receive the request directly.
LLMs receive a text version through a fixed prompt for each model family (Appendix~\ref{app:prompt}).
Section~\ref{sec:setup} explains how we obtain the option probabilities $p$ from each model.

\subsection{Non-agentic decisions}
\label{sec:static}

This component contains \nQuestionsThree{} questions across 13 domains and \nLeavesThree{} tasks (Table~\ref{tab:domains}).
Each task covers one type of decision and draws on one or more public sources.
Examples include deciding whether a message is phishing or whether a patch fixes an issue.
The domains cover text classification, knowledge, professional tasks, retrieval, structured data, code, and forecasting.
They also cover dialogue, batch processing, LLM-as-judge, safety, routing, and subjective judgments.
Appendix~\ref{app:sources} lists the sources for all benchmark tasks.

\paragraph{Labels and instructions.}
All labels come from the original sources; we do not create new labels.
They include test results, solver outputs, resolved events, expert annotations, annotator distributions, and original dataset labels.
When a source provides instructions or annotation guidelines, we preserve their wording or adapt them to the decision-request format.
Otherwise, we write instructions based on the source's labeling criteria.
This keeps the instructions consistent with how the labels were assigned.
Appendix~\ref{app:details} describes the procedure.

\subsection{Agentic tasks and environments}
\label{sec:loop}

Agentic tasks require a model to choose a sequence of actions to achieve a goal.
Each action changes the state observed at the next step, so success depends on decisions throughout an episode.
We include \nEnvs{} environments from \nEnvFamilies{} families and \nEnvSuites{} benchmark suites.
They cover web shopping, customer service, coding, robotics, games, and driving (Appendix~\ref{app:envs}).

To support System-1 models on long-horizon tasks, we use a fixed, task-specific harness.
At each step, it converts the current state and interaction history into text and presents the available actions as typed questions.
It executes the highest-probability option for each question and obtains the next observation.
We represent free-form outputs, such as search terms, tool arguments, and patches, as candidate choices.
Actions come from the original benchmark's action space, and all models use the same harness and environment.
For environments with a success criterion, we report the mean success rate, giving each environment equal weight.

\paragraph{Single-step tasks.}
We separately evaluate individual agentic decisions using recorded trajectories and fixed states from games and embodied tasks.
Each example provides a fixed state and interaction history.
Questions ask which action to take, whether a proposed step is correct, or whether the task is complete.
We score the answer against a reference without continuing the interaction.
References come from expert demonstrations, execution checks, solvers, recorded outcomes, or source annotations.
These tasks measure decision quality for a given state and history.

The full single-step collection contains 13,763 questions across 18 tasks.
It includes 6,943 questions about agent steps, 3,828 about games, and 2,992 about embodied and robotics tasks (Table~\ref{tab:domains}).
Together with the non-agentic component, these tasks account for all \nQuestions{} questions and \nLeaves{} tasks.
Our single-step model comparison uses only the agent-step subset (Table~\ref{tab:agent}).
This subset contains 3,481 recorded steps and 6,943 questions across seven tasks.
The static game and embodied tasks remain part of the benchmark but are excluded from this comparison.

\paragraph{Harness difficulty check.}
We check whether the harness makes the original task too easy.
We flag environments where a uniform-random policy succeeds in at least 20\% of episodes or models generally achieve at least 90\% success.
For flagged environments, we removed assistance or reduced the decision budget.
For example, we limited the budget to twice the shortest solution length in BabyAI, MiniHack mazes, and Baba Is AI.
A scripted policy confirms that each adjusted task remains solvable.

\paragraph{Real-time mode.}
In Atari, Doom, driving, and quadrotor environments, the world continues to evolve while the model decides.
For each decision, the environment advances by the model's measured median latency and continues to execute its previous action.
Slower models therefore act on older observations.
This makes response time part of task performance (Appendix~\ref{app:envs}).

\subsection{Metrics}
\label{sec:metrics}

We evaluate the probability distribution returned for each question.
\textbf{Accuracy} compares the option with the highest predicted probability against the hard reference label.
The \textbf{Brier score} \citep{brier1950verification} measures probability quality as $\sum_o (p(o) - t(o))^2$.
Here, $t$ is a one-hot label or an annotator distribution.
The score ranges from 0 to 2, with lower values indicating better probability estimates.

For each task, we average scores across its sources.
A domain score averages the scores of its tasks.
The overall score averages all tasks, giving each task equal weight.
Appendix~\ref{app:details} defines the additional metrics: expected calibration error, NDCG, and pair accuracy.

\section{Experimental Setup}
\label{sec:setup}

\paragraph{Probabilities from each model family.}
For a question $q$ with options $o_1, \dots, o_K$, every model provides a distribution $p(o_k \mid s, q)$.
\emph{System-1 models} return this distribution directly.
We use the APIs for Jev, Microsoft-Decision-1 and the OpenAI Decisions API, and the official code for the other System-1 models.

\emph{Closed LLMs} return a nonnegative value $v_k$ for each option in a JSON response (Appendix~\ref{app:prompt-closed}).
We normalize these values as $p(o_k \mid s, q) = v_k / \sum_j v_j$.
Missing options receive $v_k = 0$.
We do not use token probabilities from these models.

For \emph{open LLMs}, we obtain probabilities from answer labels.
The prompt $x$ contains the state, one question, and its options, each with a short label $\ell_k$.
Labels use A, B, \dots{} for choices, 0--9 for score levels, and yes/no for boolean questions (Appendix~\ref{app:prompt-open}).
The prompt ends at the start of the answer.
One forward pass gives $z_k = \log P_\theta(\ell_k \mid x)$ for each label at this position.
We normalize these values as
$
p(o_k \mid s, q)
= \frac{\exp z_k}{\sum_{j=1}^{K} \exp z_j}.
$
GPT-OSS requires reasoning, so we read label probabilities at the start of its final answer.
For DiffusionGemma, one read-only denoising step provides the label probabilities at the answer slot \citep{vllm2026structured}.
The other open models use their non-thinking mode (Appendix~\ref{app:details}).

\paragraph{Models.}
We evaluate 30 models in three families (Tables~\ref{tab:main} and~\ref{tab:agent}).
\begin{itemize}[leftmargin=*,itemsep=1pt,topsep=2pt]
  \item \textbf{Closed LLMs} (7 models, from most to least accurate):
  Claude Opus~5.5 \citep{anthropic2026opus55},
  GPT-6 Astra \citep{openai2026astra},
  Grok~4.7 \citep{xai2026grok47},
  GPT-6 Sol, GPT-5.6 Sol, GPT-6 Luna, and GPT-5.6 Luna \citep{openai2026gpt6solluna,openai2026gpt56}.
  Each model returns option probabilities in a JSON response.

  \item \textbf{Open LLMs} (13 models):
  models from the Qwen3 \citep{yang2025qwen3},
  Qwen3.5 \citep{qwen2026qwen35},
  Qwen3.8 \citep{qwen2026qwen38},
  Gemma~4 \citep{google2026gemma4},
  Llama~3 \citep{grattafiori2024llama3},
  and GPT-OSS \citep{openai2025gptoss} families,
  plus DiffusionGemma \citep{diffusiongemma2026}.

  \item \textbf{System-1 models} (10 models):
  Jev, Laya \citep{convai2026laya}, CLM \citep{kwok2026clm},
  five Decision~2.0 models \citep{vllmsr2026decision2},
  Microsoft-Decision-1 \citep{microsoft2026decision1}, which is post-trained from Qwen3.5-9B,
  and the OpenAI Decisions API~\citep{openai2026decisions}, which serves GPT-6 Luna.
\end{itemize}

We disable reasoning whenever the model's interface permits it.
Five models require reasoning: Claude Opus~5.5, GPT-6 Astra, Grok~4.7, gpt-oss-20b, and gpt-oss-120b.
We use the lowest available reasoning effort for these models.

\paragraph{Non-agentic and agentic tasks.}
The benchmark organizes its domains into four groups (Table~\ref{tab:domains}, Appendix~\ref{app:sources}).
\emph{Text and knowledge} covers text classification, knowledge and reasoning, and subjective judgments.
\emph{Applied decisions} covers professional domains, retrieval, structured data, code, forecasting, dialogue, and batch processing.
\emph{Evaluation and control} covers LLM-as-judge, safety, and routing.
\emph{Agent and action} covers decisions made during an agent's interaction with an environment.

The first three groups form the \emph{non-agentic} component, with \nQuestionsThree{} questions across 87 tasks.
These tasks ask a model to judge a given input.
The agent and action group forms the \emph{agentic} component.
We evaluate this component in \nEnvs{} environments and on 3,481 agent steps.
The recorded steps contain 6,943 questions across seven tasks. The static model comparison therefore covers 85,305 questions across 94 tasks.
The remaining 6,820 questions come from static game and embodied tasks.

\paragraph{Protocol.}
Questions with failed calls, unusable responses, or refused inputs receive no accuracy credit.
In environments, the harness executes the fallback action when a decision is missing or invalid.
Latency is the median time per request.
We report cost per 1,000 requests in each model family's native unit.
Costs use US dollars for Jev, Microsoft-Decision-1, OpenAI Decisions API, and closed LLMs.
For local models, costs use H100-equivalent GPU-seconds.
We do not convert between these units.

Output tokens are the mean number of tokens generated per request.
Returning probabilities and reading label probabilities do not themselves generate output tokens.
Appendix~\ref{app:details} describes the services, hardware, and retry rules.

\section{Results}
\label{sec:results}

\begin{table*}[t]
\centering
\caption{\textbf{All models on the non-agentic tasks: accuracy (\%), overall accuracy and Brier score,
latency, cost and output tokens.} Rank is by overall accuracy; within a family, the rows follow it. Every cell is the
mean over the tasks of a domain on all 78,362 questions. The grey columns are the overall scores. Blue rows are the three most accurate models, green rows the top three with 0 output tokens (darker is better); bold is the best per column. $^{\ast}$The model cannot stop reasoning and uses its lowest reasoning effort. Section~\ref{sec:setup} defines
latency (median per request), cost (per 1,000 requests, in USD for hosted models and H100-equivalent GPU-seconds for local models) and output
tokens (mean per request). Domains: Cls = text classification, Know = knowledge and reasoning, Social = subjective
judgments, Pro = professional domains, Ret = retrieval, Struct = structured data, Fore = forecasting, Dialog = dialogue,
Batch = batch processing, Judge = LLM-as-judge, Safety = safety and moderation, Route = routing.}
\scriptsize
\setlength{\tabcolsep}{1.5pt}
\begin{adjustbox}{center}
\resizebox{\textwidth}{!}{%
\begin{tabular}{rlccccccccccccc>{\columncolor{overallcol}}c>{\columncolor{overallcol}}ccrr}
\toprule
 & & \multicolumn{3}{c}{Text and knowledge} & \multicolumn{7}{c}{Applied decisions} & \multicolumn{3}{c}{Evaluation and control} & \multicolumn{2}{c}{All 87 tasks} & Latency & Cost per & Output \\
\cmidrule(lr){3-5} \cmidrule(lr){6-12} \cmidrule(lr){13-15} \cmidrule(lr){16-17}
Rank & Model & Cls & Know & Social & Pro & Ret & Struct & Code & Fore & Dialog & Batch & Judge & Safety & Route & Acc. & Brier & p50 (ms) & 1,000 req. & tokens \\
\midrule
\rowcolor{gray!12} \multicolumn{20}{l}{\emph{System-1 models, native probabilities}} \\
\rowcolor{zerotwo} 7 & Decision-2.0-Vega-27B & 75.1 & 80.0 & 68.0 & 70.0 & 76.7 & 83.0 & 68.9 & 43.8 & 73.5 & 77.1 & 68.4 & 85.1 & 66.5 & 73.0 & 0.343 & 228 & 316 GPU-s & 0 \\
\rowcolor{zerothree} 8 & Jev 1.13 & 73.0 & 81.9 & 69.7 & 70.6 & 77.0 & 78.7 & 65.9 & 44.1 & 73.2 & 75.3 & 68.4 & 86.7 & 64.5 & 72.6 & 0.349 & 321 & \$0.052 & 0 \\
11 & Microsoft-Decision-1 & 72.9 & 80.5 & 60.5 & 67.6 & 75.9 & 78.9 & 64.5 & 40.7 & 71.9 & 73.5 & 68.5 & 82.9 & 63.8 & 70.9 & 0.369 & 304 & \$0.038 & 0 \\
17 & OpenAI Decisions API & 69.8 & 74.5 & 38.3 & 69.1 & 73.2 & 76.0 & 66.0 & 48.3 & 70.6 & 72.2 & 67.3 & 81.0 & 65.0 & 68.6 & 0.424 & 140 & \$0.111 & 0 \\
18 & Decision-2.0-Lux-9B & 69.3 & 76.2 & 61.0 & 65.1 & 74.7 & 72.0 & 62.0 & 39.8 & 70.7 & 75.0 & 60.6 & 75.5 & 58.4 & 67.0 & 0.397 & 104 & 144 GPU-s & 0 \\
20 & Decision-2.0-Nox-4B & 67.3 & 73.6 & 58.6 & 62.2 & 72.8 & 69.4 & 58.8 & 40.1 & 70.7 & 74.4 & 58.3 & 73.8 & 59.0 & 65.1 & 0.414 & 114 & 154 GPU-s & 0 \\
26 & Decision-2.0-Kai-0.6B & 54.7 & 51.1 & 30.0 & 49.8 & 59.8 & 49.9 & 48.6 & 34.3 & 53.6 & 64.6 & 47.9 & 57.4 & 46.0 & 50.3 & 0.537 & 72 & 93 GPU-s & 0 \\
27 & Laya (421M) & 47.2 & 45.9 & 45.4 & 38.7 & 42.2 & 39.6 & 44.7 & 25.4 & 47.7 & 52.3 & 41.7 & 60.9 & 42.4 & 44.5 & 0.660 & 136 & 12 GPU-s & 0 \\
28 & CLM-v0.1-8B & 35.6 & 48.3 & 22.1 & 28.2 & 45.9 & 40.7 & 40.1 & 35.5 & 40.8 & 42.2 & 46.0 & 54.7 & 49.6 & 41.4 & 0.749 & 41 & 19 GPU-s & 0 \\
\midrule
\rowcolor{gray!12} \multicolumn{20}{l}{\emph{Closed LLMs, verbalized probabilities}} \\
\rowcolor{topone} 1 & Claude Opus 5.5$^{\ast}$ & 79.7 & \textbf{91.6} & \textbf{79.6} & \textbf{79.5} & 78.8 & \textbf{95.2} & \textbf{82.0} & \textbf{61.6} & 76.5 & \textbf{77.7} & \textbf{77.8} & 86.2 & \textbf{72.3} & \textbf{81.1} & \textbf{0.232} & 2,293 & \$10.380 & 132 \\
\rowcolor{toptwo} 2 & GPT-6 Astra$^{\ast}$ & \textbf{79.8} & 90.9 & 76.4 & 77.9 & \textbf{79.3} & 92.1 & 79.0 & 57.6 & \textbf{77.4} & 76.2 & 77.0 & \textbf{90.9} & 68.0 & 80.1 & 0.282 & 2,189 & \$15.050 & 55 \\
\rowcolor{topthree} 3 & Grok 4.7$^{\ast}$ & 76.4 & 88.1 & 73.1 & 73.4 & 77.7 & 88.8 & 76.0 & 52.1 & 74.4 & 72.9 & 75.0 & 89.6 & 70.6 & 77.5 & 0.306 & 6,694 & \$7.569 & 765 \\
4 & GPT-6 Sol & 78.0 & 84.6 & 74.5 & 74.8 & 77.6 & 87.9 & 74.2 & 53.3 & 75.7 & 74.9 & 74.8 & 90.7 & 69.9 & 77.4 & 0.340 & 1,446 & \$2.833 & 37 \\
5 & GPT-5.6 Sol & 77.1 & 83.2 & 73.2 & 75.0 & 77.9 & 86.3 & 72.9 & 51.2 & 75.3 & 74.9 & 74.0 & 90.9 & 69.8 & 76.7 & 0.349 & 1,680 & \$5.817 & 45 \\
10 & GPT-6 Luna & 72.4 & 75.3 & 69.0 & 71.3 & 75.2 & 77.3 & 66.4 & 49.2 & 71.9 & 73.1 & 67.1 & 86.5 & 62.2 & 71.2 & 0.461 & 1,276 & \$0.143 & 40 \\
13 & GPT-5.6 Luna & 71.2 & 72.5 & 63.0 & 70.5 & 74.8 & 76.0 & 64.7 & 46.0 & 72.6 & 73.9 & 67.6 & 85.6 & 62.3 & 70.0 & 0.458 & 1,375 & \$0.301 & 46 \\
\midrule
\rowcolor{gray!12} \multicolumn{20}{l}{\emph{Open LLMs, probabilities read at the answer token}} \\
\rowcolor{zeroone} 6 & Gemma-4-31B & 73.7 & 80.8 & 65.8 & 72.1 & 78.0 & 82.6 & 69.4 & 42.7 & 72.4 & 74.5 & 71.7 & 88.7 & 70.0 & 73.9 & 0.474 & 60 & 142 GPU-s & 0 \\
9 & gpt-oss-120b$^{\ast}$ & 69.5 & 82.9 & 57.9 & 67.9 & 74.5 & 82.9 & 71.6 & 40.9 & 72.3 & 71.3 & 70.7 & 82.8 & 64.3 & 71.8 & 0.534 & 284 & 101 GPU-s & 95 \\
12 & Qwen3.8-27B & 71.9 & 75.5 & 60.7 & 69.8 & 74.0 & 72.5 & 65.1 & 40.1 & 69.7 & 74.6 & 67.9 & 86.3 & 66.4 & 70.2 & 0.391 & 105 & 142 GPU-s & 0 \\
14 & Gemma-4-26B-A4B & 71.8 & 76.2 & 51.2 & 68.3 & 75.6 & 77.9 & 64.4 & 40.9 & 71.4 & 75.3 & 67.2 & 83.7 & 66.5 & 69.9 & 0.528 & 47 & 25 GPU-s & 0 \\
15 & DiffusionGemma-26B-A4B & 71.2 & 74.7 & 68.2 & 68.0 & 76.2 & 75.5 & 63.0 & 42.1 & 69.9 & 73.8 & 66.0 & 83.6 & 63.8 & 69.7 & 0.465 & 68 & 41 GPU-s & 0 \\
16 & gpt-oss-20b$^{\ast}$ & 67.2 & 79.8 & 48.9 & 64.2 & 71.4 & 79.0 & 70.0 & 41.6 & 69.5 & 68.7 & 67.0 & 80.6 & 61.3 & 68.7 & 0.589 & 183 & 51 GPU-s & 109 \\
19 & Qwen3.5-9B & 67.5 & 72.0 & 57.4 & 65.5 & 69.3 & 68.6 & 60.5 & 37.8 & 60.0 & 70.2 & 63.9 & 77.7 & 60.4 & 65.3 & 0.461 & 77 & 36 GPU-s & 0 \\
21 & Qwen3-32B & 63.1 & 72.9 & 55.8 & 64.4 & 71.4 & 67.4 & 58.0 & 40.7 & 69.8 & 72.1 & 63.4 & 75.8 & 58.1 & 65.0 & 0.543 & 50 & 112 GPU-s & 0 \\
22 & Llama-3.3-70B & 64.1 & 71.9 & 55.4 & 64.0 & 71.3 & 66.4 & 58.8 & 39.9 & 67.1 & 73.3 & 59.4 & 77.7 & 63.2 & 64.8 & 0.640 & 46 & 241 GPU-s & 0 \\
23 & Gemma-4-E4B & 64.4 & 65.9 & 58.0 & 61.6 & 69.3 & 64.0 & 56.4 & 38.6 & 65.2 & 72.6 & 59.0 & 75.4 & 60.5 & 62.8 & 0.602 & 29 & 21 GPU-s & 0 \\
24 & Qwen3-8B & 56.6 & 65.8 & 46.8 & 59.2 & 64.5 & 63.5 & 54.8 & 39.4 & 57.7 & 68.7 & 59.3 & 69.5 & 54.0 & 59.3 & 0.715 & 30 & 27 GPU-s & 0 \\
25 & Llama-3.1-8B & 52.2 & 56.8 & 50.1 & 51.8 & 55.1 & 53.8 & 48.4 & 37.3 & 53.2 & 63.1 & 50.3 & 66.5 & 50.1 & 53.1 & 0.645 & 21 & 25 GPU-s & 0 \\
29 & Qwen3-0.6B & 24.3 & 44.8 & 36.7 & 34.7 & 41.2 & 47.5 & 42.5 & 24.8 & 39.6 & 41.5 & 46.7 & 42.2 & 39.1 & 39.2 & 0.937 & 27 & 8 GPU-s & 0 \\
\bottomrule
\end{tabular}
}
\end{adjustbox}
\label{tab:main}
\end{table*}

\begin{table}[t]
\centering
\caption{\textbf{The agentic tasks (\%).} Rank is by step accuracy; within a family, the rows follow it.
``Single steps'': accuracy on 3,481 agent steps (mean over 7 tasks) and the share of their 6,943 questions without an answer.
``Closed loops'': success rate over 127 environments and the share of decisions without an answer.
Blue and green mark the top three models by step accuracy overall and with 0 output tokens, respectively;
green takes precedence for models in both groups (darker is better). Bold marks the best value in a column.
$^{\ast}$Reasons at its lowest effort. ``No ans.'' is the percentage of questions (steps) or decisions (loops) without an answer.}
\scriptsize
\setlength{\tabcolsep}{3pt}
\begin{adjustbox}{center}
\resizebox{\linewidth}{!}{%
\begin{tabular}{rlcccc}
\toprule
 & & \multicolumn{2}{c}{Single steps} & \multicolumn{2}{c}{Closed loops} \\
\cmidrule(lr){3-4} \cmidrule(lr){5-6}
Rank & Model & Acc. & No ans. & Success & No ans. \\
\midrule
\rowcolor{gray!12} \multicolumn{6}{l}{\emph{System-1 models, native probabilities}} \\
7 & Jev 1.13 & 67.8 & 13.2 & 17.4 & 0.4 \\
8 & Decision-2.0-Vega-27B & 67.8 & 13.7 & 26.5 & 0.0 \\
10 & OpenAI Decisions API & 67.2 & 0.1 & 16.9 & 0.7 \\
13 & Microsoft-Decision-1 & 65.2 & 13.7 & 12.4 & 0.0 \\
16 & Decision-2.0-Sol-2B & 46.4 & 24.7 & 3.1 & 0.0 \\
18 & Laya (421M) & 41.4 & 0.1 & 1.3 & 0.8 \\
19 & CLM-v0.1-8B & 40.6 & 0.0 & 1.7 & 0.0 \\
20 & Decision-2.0-Kai-0.6B & 34.2 & 40.9 & 2.2 & 7.2 \\
\midrule
\rowcolor{gray!12} \multicolumn{6}{l}{\emph{Closed LLMs, verbalized probabilities}} \\
\rowcolor{toptwo} 2 & GPT-6 Luna & 69.1 & 0.1 & 22.9 & 0.2 \\
9 & GPT-5.6 Luna & 67.4 & 0.7 & 18.9 & 0.1 \\
\midrule
\rowcolor{gray!12} \multicolumn{6}{l}{\emph{Open LLMs, probabilities read at the answer token}} \\
\rowcolor{zeroone} 1 & Gemma-4-31B & \textbf{71.3} & 0.3 & \textbf{32.9} & 0.0 \\
\rowcolor{zerotwo} 3 & Qwen3.8-27B & 68.9 & 0.0 & 26.2 & 0.0 \\
\rowcolor{zerothree} 4 & DiffusionGemma-26B-A4B & 68.2 & 0.0 & 13.0 & 0.0 \\
5 & Gemma-4-26B-A4B & 68.1 & 0.0 & 21.7 & 0.0 \\
6 & Qwen3.5-9B & 67.8 & 0.0 & 5.7 & 0.0 \\
11 & Gemma-4-E4B & 65.9 & 0.3 & 5.1 & 0.0 \\
12 & gpt-oss-20b$^{\ast}$ & 65.8 & 0.3 & 24.7 & 0.0 \\
14 & Qwen3-32B & 62.8 & 10.2 & 6.6 & 0.0 \\
15 & Qwen3-8B & 59.9 & 10.2 & 2.5 & 0.0 \\
17 & Qwen3-0.6B & 43.9 & 10.2 & 1.8 & 0.0 \\
\bottomrule
\end{tabular}
}
\end{adjustbox}
\label{tab:agent}
\end{table}

\begin{figure}[t]
\centering
\begin{minipage}[t]{0.43\linewidth}
  \vspace{0pt}
  \centering
  \footnotesize
  \renewcommand{\arraystretch}{1.05}
  \setlength{\tabcolsep}{3pt}
  \resizebox{\linewidth}{!}{%
    \begin{tabular}{@{}lr@{}}
      \toprule
      \textbf{Task domain} & \textbf{Gap (pp)} \\
      \midrule
      Filter, group \& rank & 0.5 \\
      Retrieval \& ranking & 2.3 \\
      Dialogue decisions & 3.9 \\
      Safety \& moderation & 4.2 \\
      Text classification & 4.7 \\
      Model \& tool routing & 5.8 \\
      Professional tasks & 8.9 \\
      LLM evaluation & 9.3 \\
      Knowledge \& reasoning & 9.7 \\
      Social judgments & 9.9 \\
      Structured data & 12.2 \\
      Code understanding & 13.1 \\
      Event forecasting & 13.3 \\
      \bottomrule
    \end{tabular}%
  }
\end{minipage}\hfill%
\begin{minipage}[t]{0.57\linewidth}
  \vspace{0pt}
  \centering
  \includegraphics[width=\linewidth]{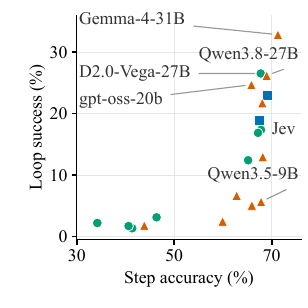}
\end{minipage}

\caption{\textbf{Task-domain gaps and agentic performance.}
Left: gaps by task domain in percentage points.
Right: agentic task performance.
Evaluation settings are given in Section~\ref{sec:setup}.}
\label{fig:front-agentic}
\end{figure}

Table~\ref{tab:main} reports non-agentic performance: accuracy across 13 domains, overall accuracy, Brier score, latency, cost, and output tokens.
Table~\ref{tab:agent} reports agentic performance.

\paragraph{Non-agentic: closed LLMs with reasoning are the most accurate.}
Claude Opus~5.5 achieves the highest overall accuracy (81.1\%) and the lowest Brier score, leading in 9 of the 13 domains.
The five most accurate models are closed LLMs, three of which reason before answering.
Gemma-4-31B is the most accurate open LLM (73.9\%).
Without reasoning, it outperforms both Luna models in all three non-agentic groups.
Forecasting is the hardest domain for 25 of the 29 models with non-agentic results, with no model exceeding 61.6\% accuracy.

\paragraph{Non-agentic: the best System-1 models match low-reasoning-effort LLMs.}
The best $\text{System-1}$ models match or exceed low-reasoning-effort LLMs in accuracy on non-agentic tasks.
Decision-2.0-Vega-27B (73.0\%) and Jev (72.6\%) outperform GPT-6 Luna (71.2\%) and GPT-5.6 Luna~(70.0\%).
Both remain within 1.4 percentage points of Gemma-4-31B.

\paragraph{Non-agentic: System-1 models provide reliable probabilities.}
Across the 87 tasks, Decision-2.0-Vega-27B and Jev achieve mean Brier scores of 0.343 and 0.349, respectively.
Only four models achieve lower scores, all closed LLMs that cost 20--105 times as much as GPT-6 Luna.
Although Gemma-4-31B is more accurate than these two System-1 models, its Brier score is higher at 0.474.
GPT-6 Luna and GPT-5.6 Luna also have higher Brier scores, at 0.461 and 0.458, respectively.
Answer-token probabilities can be overconfident: Gemma-4-31B assigns a probability of at least 0.999 to its option on 87\% of questions.

\begin{figure}[t]
\centering
\includegraphics[width=\linewidth]{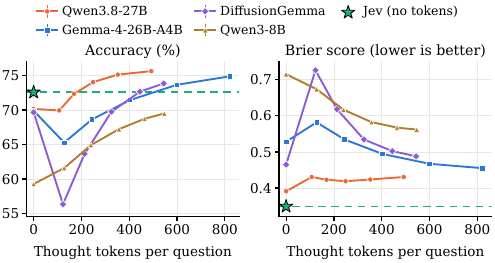}
\caption{\textbf{Thinking can improve accuracy while worsening Brier scores.}
Accuracy and Brier score on non-agentic tasks versus the mean number of reasoning tokens per question.
The zero-token point disables thinking, as in Table~\ref{tab:main}.
Other points use reasoning token limits from 128 to 2,048.}
\label{fig:think}
\end{figure}
\paragraph{Non-agentic: thinking improves accuracy, but not necessarily calibration.}
Figure~\ref{fig:think} evaluates four open LLMs with reasoning budgets of up to 2,048 tokens.
Without thinking, all four are less accurate than Jev (59.3--70.2\% versus 72.6\%).
Small reasoning budgets can reduce accuracy below this baseline.
At 128 tokens, DiffusionGemma drops from 69.7\% to 56.3\%, the largest decline among the four models, while Gemma-4-26B-A4B loses 4.6 percentage points.
Larger budgets improve accuracy for all four models, and three surpass Jev at 2,048 tokens.
However, these gains do not necessarily improve calibration.
For example, a 2,048-token reasoning budget raises Qwen3.8-27B's accuracy from 70.2\% to 75.6\%, but worsens its Brier score from 0.391 to 0.430.
Its mean confidence in the selected option rises from 81.4\% to 97.4\%.
This widening gap is consistent with increased overconfidence: predictions become more concentrated on a single option, including when that option is wrong.
Confident errors incur larger Brier penalties, so higher accuracy can coexist with less reliable probability estimates.

\paragraph{Non-agentic: forecasting remains challenging for System-1 models.}
Despite competitive overall performance, System-1 models still lag substantially in some domains, particularly forecasting.
One task, for example, asks which of three energy commodities---West Texas Intermediate crude oil, Brent crude oil, or Henry Hub natural gas---will have the largest percentage price increase over a 90-day horizon.
Across the forecasting domain, the best System-1 model, the OpenAI Decisions API, achieves 48.3\% accuracy, compared with 61.6\% for Claude Opus~5.5, a gap of 13.3 percentage points~(Table~\ref{tab:main}).
Strong performance on classification and selection tasks therefore does not necessarily extend to forecasting.
Although their low latency is attractive for applications such as trading, using System-1 models as forecasting components requires careful design and validation.

\paragraph{Agentic: similar step accuracy can yield different closed-loop success rates.}
Model rankings differ in agentic tasks (Table~\ref{tab:agent}).
Nine models achieve step accuracies of 67.4--71.3\%, but their closed-loop success rates range from 5.7\% to 32.9\%.
Decision-2.0-Vega-27B, Jev, and Qwen3.5-9B all achieve 67.8\% step accuracy, yet their closed-loop success rates are 26.5\%, 17.4\%, and 5.7\%, respectively.
One possible explanation is that errors accumulate during interaction.
A static step starts from a recorded state, whereas an incorrect closed-loop decision can alter all subsequent states.

\paragraph{Agentic: larger models achieve higher closed-loop success rates.}
Gemma-4-31B achieves the highest closed-loop success rate (32.9\%), followed by Decision-2.0-Vega-27B (26.5\%) and Qwen3.8-27B (26.2\%).
The four open LLMs with fewer than 10B parameters succeed in fewer than 6\% of episodes.
Qwen3-32B is an exception among the larger models, with a success rate of only 6.6\%.
Jev achieves 17.4\%, below both Luna models (22.9\% and 18.9\%) but above six of the ten open LLMs.

\begin{figure}[t]
\centering
\begin{minipage}[c]{0.55\linewidth}
  \centering
  \resizebox{\linewidth}{!}{\begin{tabular}{@{}lrrrr@{}}
\toprule
 & \multicolumn{2}{c}{Non-agentic} & \multicolumn{2}{c}{Agentic} \\
\cmidrule(lr){2-3} \cmidrule(l){4-5}
System-1 model & Acc. & Brier & Step & Loop \\
\midrule
D2.0-Vega-27B & \textcolor{better}{$+$2.9} & \textcolor{better}{$-$.048} & \textcolor{worse}{$-$1.2} & $+$0.3 \\
\quad\textcolor{gray}{from Qwen3.8-27B} & & & & \\[1pt]
MS-Decision-1 & \textcolor{better}{$+$5.6} & \textcolor{better}{$-$.091} & \textcolor{worse}{$-$2.7} & \textcolor{better}{$+$6.8} \\
\quad\textcolor{gray}{from Qwen3.5-9B} & & & & \\[1pt]
D2.0-Kai-0.6B & \textcolor{better}{$+$11.0} & \textcolor{better}{$-$.400} & \textcolor{worse}{$-$9.6} & $+$0.4 \\
\quad\textcolor{gray}{from Qwen3-0.6B$^{\dagger}$} & & & & \\[1pt]
CLM-v0.1-8B & \textcolor{worse}{$-$17.8} & \textcolor{worse}{$+$.034} & \textcolor{worse}{$-$19.3} & $-$0.8 \\
\quad\textcolor{gray}{from Qwen3-8B} & & & & \\[1pt]
\midrule
Decisions API & \textcolor{worse}{$-$2.6} & \textcolor{better}{$-$.037} & \textcolor{worse}{$-$2.0} & \textcolor{worse}{$-$6.1} \\
\quad\textcolor{gray}{from GPT-6 Luna$^{\ddagger}$} & & & & \\[1pt]
\bottomrule
\end{tabular}
}
\end{minipage}\hfill%
\begin{minipage}[c]{0.44\linewidth}
  \centering
  \includegraphics[width=\linewidth]{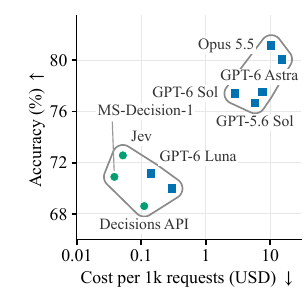}
\end{minipage}
\caption{\textbf{Paired comparisons show mixed gains in non-agentic accuracy but consistently lower agentic step accuracy (left); higher cost does not always yield higher accuracy (right).}
$^{\dagger}$Base: Qwen3-0.6B-Base.
$^{\ddagger}$Same LLM, different interface.}
\label{fig:cost-accuracy}
\end{figure}

\paragraph{Paired comparisons: System-1 models versus their corresponding LLMs.}
We compare five System-1 models with their corresponding LLMs, pairing models that differ in their decision interface or post-training.
System-1 models improve non-agentic accuracy in three pairs but reduce agentic step accuracy in all five.
These comparisons include four post-trained models and the Decisions API, which uses the same underlying LLM as GPT-6 Luna (Figure~\ref{fig:cost-accuracy}, left).
On non-agentic tasks, Decision-2.0-Vega-27B, $\text{Microsoft-Decision-1}$, and Decision-2.0-Kai-0.6B improve accuracy by 2.9--11.0 percentage points.
The Decisions API and CLM-v0.1-8B instead lose 2.6 and 17.8 percentage points, respectively.
Brier scores improve in four of the five pairs, with reductions of 0.037--0.400; CLM is the only exception.
On agentic tasks, step accuracy decreases by 1.2--19.3 percentage points across all five pairs.
Closed-loop success changes little in three pairs, staying within 0.8 percentage points of the corresponding LLM.
It decreases from 22.9\% to 16.9\% for the Decisions API but increases from 5.7\% to 12.4\% for Microsoft-Decision-1.

\paragraph{Latency.}
Local models return decisions fastest.
Without reasoning, open LLMs take 21--105~ms per request, while local System-1 models take 41--228~ms.
The two GPT-OSS models use reasoning and take 183 and 284~ms.
Jev takes 321~ms through the gateway and Microsoft-Decision-1 takes 304~ms, while closed LLMs take 1.3--6.7~s.

\paragraph{Cost and output tokens.}
Higher cost does not always yield higher accuracy among hosted models (Figure~\ref{fig:cost-accuracy}, right).
Jev achieves 72.6\% accuracy at \$0.052 per 1,000 requests, compared with 71.2\% at \$0.14 for GPT-6 Luna.
Microsoft-Decision-1 is the least expensive hosted model, achieving 70.9\% at \$0.038.
Claude Opus~5.5 achieves 81.1\% at \$10.38, roughly 74 times GPT-6 Luna's cost for a 9.9-percentage-point accuracy gain.
GPT-6 Astra is the most expensive (\$15.05) but is less accurate than Opus (80.1\% versus 81.1\%).
Output length also varies substantially: Grok~4.7 generates 765 tokens per request, Claude Opus~5.5 generates 132, and GPT-6 Luna generates 40.
Jev generates no text.
Among local models, Gemma-4-26B-A4B uses 25 GPU-seconds per 1,000 requests, compared with 142 for Gemma-4-31B, which achieves 4.0 percentage points higher overall accuracy.

\section{Conclusion}
\label{sec:conclusion}

\name{} asks one question of every decision: can it be left to a fast model that answers with calibrated probabilities in a
single forward pass, or does it still need a generative LLM? Posing the same typed requests to 30 models makes the question measurable. The answers are encouraging for System~1 on most static
decisions, where the largest System-1 models match fast closed LLMs with better-calibrated probabilities at a quarter of the
latency, and sobering where the best accuracy is needed, which a reasoning LLM delivers at about 70 times the cost. In closed
loops, large models succeed most often, step accuracy does not predict loop success, and small System-1
models lack the planning that a loop needs. We hope the benchmark helps decide, decision by decision, where deliberation is worth its price.

\section{Limitations}
\label{sec:limitations}

Labels are inherited from the sources, so their noise is inherited too; tier-C
instructions were written by us; and many sources are public and several years old, so some inputs may have been seen in
training. The closed LLMs, Jev, Microsoft-Decision-1 and the OpenAI Decisions API were called through online services, so their latency includes
the service and the network, and cost reflects the prices on the days of the runs. The models that sample their answers
got one draw per request, so run-to-run variation is not measured.
\section{Ethical Considerations}
\label{sec:ethics}

\name{} reuses public datasets and environments under their licenses (Table~\ref{tab:sources}); sources whose terms do
not allow redistribution are rebuilt locally by a loader and are not redistributed. Toxicity, hate-speech, and safety
tasks contain offensive or unsafe text by design; we keep these inputs as their creators released them and add no new
personal data. A high score on the benchmark does not show that a model is fit for decisions with legal, medical,
financial, or safety consequences, where errors can harm people and human review remains necessary.

\bibliography{references,datasets}  %

\clearpage \newpage

\appendix
\section{LLM Prompts}
\label{app:prompt}

The closed LLMs and the open LLMs get different prompts, because we get their probabilities in different ways
(Section~\ref{sec:setup}). Each family has one fixed prompt, and no model gets a prompt that we tuned for it. Both examples
below use the same request of the date-arithmetic source (knowledge and reasoning).

\subsection{Closed LLMs: verbalized probabilities}
\label{app:prompt-closed}

A closed LLM writes its probabilities as numbers in a JSON object. We do not read the probabilities of its tokens. All closed
LLMs get the same two messages, and one request with all its questions is one call. The system message is:
\begin{quote}\small\ttfamily\raggedright
You are a classifier. Read the input and answer every question. For each question give a probability for the allowed
options; the probabilities of one question must sum to 1. Reply with a single JSON object and nothing else.
\end{quote}
The user message gives the request:
\begin{quote}\scriptsize
\begin{verbatim}
Input:
{
 "date": "1968-03-18"
}

Questions:
- "date" (choice): Which date is 80 days
  after the given date?
    - "A": June 6, 1969
    - "B": July 6, 1968
    - "C": June 5, 1968
    - "D": June 6, 1968

Reply with JSON of the form
{"<question>": {"<option id>": <probability>,
 ...}, ...}.
Options you leave out get probability 0.
\end{verbatim}
\end{quote}
(We break the two long lines here for the column width; the model gets them unbroken.) A boolean question lists the options
\texttt{"true"} and \texttt{"false"}. A score question lists its levels as \texttt{"0"}, \texttt{"1"}, \dots{} with the
description of each level.

\subsection{Open LLMs: probabilities of label tokens}
\label{app:prompt-open}

An open LLM does not write probabilities. It gets one question per prompt, and we read the probability of each label token
at the position where the answer starts (Section~\ref{sec:setup}). The chat template of the model contains the two
messages and opens the reply of the assistant; with no reasoning, the prompt ends there. The system message is:
\begin{quote}\small\ttfamily\raggedright
You are a classifier. Read the input and answer the question. Reply with the label of your answer only.
\end{quote}
The user message gives the state and one question:
\begin{quote}\scriptsize
\begin{verbatim}
Input:
{
 "date": "1968-03-18"
}

Question: Which date is 80 days after the
  given date?
Options:
A. June 6, 1969
B. July 6, 1968
C. June 5, 1968
D. June 6, 1968
Reply with the label of one option.
\end{verbatim}
\end{quote}
(We break the long line again.) We read the log-probabilities of the tokens \texttt{A}, \texttt{B}, \texttt{C} and \texttt{D}. A boolean question ends with
``Reply yes or no.'', and we read \texttt{yes} and \texttt{no} (with \texttt{Yes} and \texttt{No} added in). A score question
lists its levels as \texttt{0.}, \texttt{1.}, \dots{} and ends with ``Reply with the number of one level.''
DiffusionGemma gets the same two messages. GPT-OSS gets them in its own chat format, and we read the labels after its
reasoning message.

\subsection{System-1 models}

Jev gets the request through its native interface: the state as JSON and, for each question, its type, instructions and
option or level descriptions. The OpenAI Decisions API gets the same request: the state is the JSON text that the LLMs see, a
boolean question is a predicate, a choice question keeps its options and their descriptions, and a score question gives
each level the label 0 to $n-1$ and its description. The API returns $P(\text{true})$ for a predicate and a distribution
over the options or levels for the other types. OpenRouter's Decisions API serves Microsoft-Decision-1 and takes requests in the
same format as Jev's interface; only the type name of a boolean question differs (\texttt{noul}). The API returns $P(\text{true})$ for a boolean question and a
distribution over the options or levels for the other types. The other System-1 models get the request through the code
that their authors released.

\section{Evaluation Details}
\label{app:details}

\paragraph{Static tasks.} Each task has about 500 requests, split evenly across its sources, sampled with a fixed seed and
stratified by label; tasks that target rare events are balanced instead. Some sources produce several requests per record (a
vulnerable function and its fixed version, two orderings of a pairwise comparison); these share a group identifier for the
pair metric. \nSoft{} requests carry a soft label, such as the share of 100 annotators who chose each label. Tier~C
instructions (\pctTierC{}\% of the requests) are written by us from the source's criteria. There is no shared input budget:
every request is sent in full. The \nLoaderOnly{} requests whose licenses do not allow redistribution are rebuilt locally by
loaders.

\paragraph{Metrics.} With $k$ samples per request, accuracy is avg@$k$, and maj@$k$ takes a majority vote; this paper uses
$k=1$. The expected calibration error \citep{guo2017calibration} uses the confidence of the top option with 10 equal-mass
bins. Tasks whose requests rank candidates for a query also report NDCG of the expected score levels, and paired requests
report pair accuracy, the share of pairs with all questions correct. A question without an answer counts as the uniform
distribution for the Brier score, the expected calibration error and NDCG.

\paragraph{Reading open models.} For a boolean question, the probability of the label \texttt{yes} includes the token
\texttt{Yes}, and \texttt{no} includes \texttt{No}. A question with more than 26 options gets number labels of several tokens
(``001'' to ``151''); we score them on a prefix tree of their tokens, and $z_k$ is the sum of the token log-probabilities. We
log the probability mass on the labels to check that the answer position is correct. GPT-OSS writes a reasoning message of at
most 4,096 tokens (2,048 on the agentic tasks) at its lowest effort, and we read the labels at the first token of
its final message. For DiffusionGemma, the answer label is one slot of the denoising canvas, all other canvas positions are
fixed, and one read-only denoising step returns the label probabilities at the slot. All open models run with vLLM
\citep{kwon2023vllm}.

\paragraph{Thinking budgets.} Four open LLMs (Qwen3-8B, Qwen3.8-27B, Gemma-4-26B-A4B and DiffusionGemma) answer every
question of the non-agentic tasks in their thinking mode. Each question gets one thought, sampled with the model's own
generation settings and written up to 2,048 tokens; the model is not told its budget. For a budget $B$ of 0, 128, 256, 512,
1,024 or 2,048 tokens, we keep the first $B$ tokens of this thought (or all of it, if the model ended it sooner), close it as
the chat template closes a thought, and read the label probabilities at the next position as in Section~\ref{sec:setup}.
The budgets therefore differ only in where one sample is cut. DiffusionGemma writes its thought in blocks of 16 tokens and
is read at the first answer token. Figure~\ref{fig:think} starts each curve at thinking off, the setting of Table~\ref{tab:main}. We also ran budget 0, the
thinking prompt with an empty thought. It generates no tokens either, but its prompt differs from thinking off: Qwen3.8-27B
adds a line that sets the reasoning effort, and Gemma-4-26B-A4B and DiffusionGemma add a thinking flag and an empty thought
channel (for Qwen3-8B the two prompts are the same string). Under this prompt, the models often want to go on thinking
instead of answering, and budget 0 is less accurate than thinking off by 2.1 points (Qwen3.8-27B), 3.6 (Gemma-4-26B-A4B)
and 13.5 (DiffusionGemma). The 95\% intervals come from a paired bootstrap stratified by source. The four models used
58.8 GPU-hours on H100 GPUs for all budgets.

\paragraph{Inputs.} We do not shorten an input, with two exceptions: the Laya library cuts an input to 512 tokens, and for
CLM we keep the last 2,048 tokens of each text, the end that its training keeps.

\paragraph{Calls and episodes.} A call that fails in transport or on a rate limit is retried. A reply that cannot be parsed,
and a request that a model refuses, are final, and a model never gets a second sample. The refusals of the OpenAI Decisions API
are deterministic: the same request sent again gets the same refusal. An environment allows its own number
of decisions per episode and has 1 to 114 episodes. Of the \nEnvs{} environments, 128 report success or failure and 10 only a
score; one of the 128 (the coding controller with a GPT-4.1 mini proposer) was played only by Jev, so the loop success is a
mean over 127 environments. The static games and embodied tasks of the agent and action group are not part of this
evaluation.

\paragraph{Services and hardware.} We call the closed LLMs through GitHub Copilot for the non-agentic tasks; for the agentic
tasks, we call the two Luna models through Amazon Bedrock. Non-agentic Copilot costs use the recorded rate of \$0.01 per usage credit. Copilot charges each model its listed credits per
million input and output tokens; a GPT-5.6 model costs twice as many credits per input token as the GPT-6 model of the same
tier (400 for GPT-5.6 Sol, 200 for GPT-6 Sol). Jev runs through the Vercel AI Gateway. We call the OpenAI Decisions API
(public beta, October 2026) directly; its cost is the list price of \$0.10 for one million input tokens, and it does not
charge for output. We call Microsoft-Decision-1 through OpenRouter's Decisions API (alpha, October 2026; served by Azure);
its cost is the cost that OpenRouter reports, \$0.042 for one million input tokens, with no charge for output. The other System-1 models run locally with the code that their authors released. Local models run on H100 GPUs for the non-agentic tasks and on A100
GPUs for the agentic tasks. Latency is the median time of single requests: on an idle GPU server for a local model, and through the
network for the closed LLMs, Jev, Microsoft-Decision-1, and the OpenAI Decisions API.

\section{Domains and Sources of Static Tasks}
\label{app:sources}

Table~\ref{tab:domains} gives the size of every domain. Table~\ref{tab:sources} lists every source with its task, number of
requests, instruction tier and license. All instructions and questions are in English, and so are the inputs of almost
all sources; the exceptions are FanLang-9 and DuLeMon (Chinese), WiLI-2018 (many languages, for language
identification), and the CORD receipts (Indonesian). The source column names the dataset and, where the task cards give it, the paper
that introduced it. A BibTeX entry per source, with the dataset's page and license, is in \texttt{sources.bib} of the paper
repository; construction notes and the exact instructions of every source are in the task cards of the code repository.

\begin{table}[t]
\centering
\caption{\textbf{The static part of the benchmark.} A task is one kind of decision, and a source is one public dataset.
The static questions of the agent and action group are the single agent steps; the games and embodied domains also have
static tasks, which the full evaluation does not cover.}
\footnotesize
\setlength{\tabcolsep}{3pt}
\begin{adjustbox}{max width=\columnwidth}
\begin{tabular}{@{}lrrr@{}}
\toprule
Group and domain & Tasks & Sources & Questions \\
\midrule
\multicolumn{4}{@{}l}{\textbf{Text and knowledge}} \\
\hspace{0.6em}Basic text classification & 10 & 26 & 14,250 \\
\hspace{0.6em}Knowledge and reasoning & 12 & 30 & 6,000 \\
\hspace{0.6em}Subjective and pluralistic & 4 & 14 & 2,846 \\
\multicolumn{4}{@{}l}{\textbf{Applied decisions}} \\
\hspace{0.6em}Professional domains & 9 & 26 & 8,313 \\
\hspace{0.6em}Retrieval, ranking and matching & 5 & 13 & 2,499 \\
\hspace{0.6em}Structured data, tables and logs & 5 & 13 & 3,500 \\
\hspace{0.6em}Code & 8 & 17 & 3,886 \\
\hspace{0.6em}Forecasting & 3 & 7 & 1,420 \\
\hspace{0.6em}Dialogue and interaction & 5 & 10 & 4,722 \\
\hspace{0.6em}Batch semantic processing & 3 & 6 & 1,500 \\
\multicolumn{4}{@{}l}{\textbf{Evaluation and control}} \\
\hspace{0.6em}LLM-as-judge and evaluators & 10 & 24 & 10,469 \\
\hspace{0.6em}Safety and moderation & 6 & 16 & 9,999 \\
\hspace{0.6em}Routing and orchestration & 7 & 18 & 8,958 \\
\multicolumn{4}{@{}l}{\textbf{Agent and action}} \\
\hspace{0.6em}Agent single steps & 7 & 18 & 6,943 \\
\hspace{0.6em}Games (static) & 5 & 13 & 3,828 \\
\hspace{0.6em}Embodied and robotics (static) & 6 & 17 & 2,992 \\
\midrule
All & 105 & 268 & 92,125 \\
\bottomrule
\end{tabular}
\end{adjustbox}
\label{tab:domains}
\end{table}

\begin{table*}[p]
\centering
\scriptsize
\setlength{\tabcolsep}{3pt}
\caption{\textbf{All sources of the static tasks.} Tier A keeps the source's instruction verbatim, B adapts it, C is newly written from the source's criteria. ``(loader)'' marks sources whose requests are rebuilt locally by a loader instead of being redistributed.}
\label{tab:sources}
\begin{tabular}{@{}>{\raggedright\arraybackslash}p{2.6cm}>{\raggedright\arraybackslash}p{7.0cm}rc>{\raggedright\arraybackslash}p{3.2cm}@{}}
\toprule
Task & Source & Requests & Tier & License \\
\midrule
\multicolumn{5}{l}{\textbf{Basic text classification}} \\
entity\_\allowbreak{}relation & Few-NERD supervised, fine-grained entity types (\citealp{ding2021fewnerd}) & 167 & C & CC BY-SA 4.0 \\
entity\_\allowbreak{}relation & FewRel 1.0, the 80 public Wikidata relations (\citealp{han2018fewrel}) & 167 & A & MIT \\
entity\_\allowbreak{}relation & SemEval-2010 Task 8, relations between nominals (\citealp{hendrickx2010semeval}) & 166 & B & CC BY 3.0 \\
intent & CLINC150, plus variant with out-of-scope queries (\citealp{larson2019evaluation}) & 250 & C & CC BY 3.0 \\
intent & MASSIVE, en-US (\citealp{fitzgerald2023massive}) & 250 & C & CC BY 4.0 \\
language\_\allowbreak{}style & CEFR-SP, public part (\citealp{arase2022cefr}) & 167 & B & CC BY-SA 3.0 (Wiki-Auto part) / CC BY\dots{} \\
language\_\allowbreak{}style & Formality scores (\citealp{pavlick2016empirical}) & 167 & B & CC BY 3.0 \\
language\_\allowbreak{}style & WiLI-2018 Wikipedia language identification (\citealp{thoma2018wili}) & 166 & C & ODbL 1.0 \\
multilabel & GoEmotions (\citealp{demszky2020goemotions}) & 250 & A & Apache-2.0 \\
multilabel & SemEval-2018 Task 1 Affect in Tweets, emotion classification E-c (\citealp{mohammad2018semeval}) & 250 & A & Research use only (loader) \\
nli & Adversarial NLI, round 3 test (\citealp{nie2020adversarial}) & 250 & B & CC BY-NC 4.0 \\
nli & ChaosNLI, SNLI and MNLI parts, 100 annotations per example (Nie et al., 2020) & 250 & A & CC BY-NC 4.0 \\
sentiment & SemEval-2018 Task 1 Affect in Tweets, valence ordinal classification (\citealp{mohammad2018semeval}) & 167 & A & Research use only (loader) \\
sentiment & Stanford Sentiment Treebank, 5 classes (\citealp{socher2013recursive}) & 167 & C & Unspecified (research use) \\
sentiment & TweetEval sentiment (SemEval-2017 Task 4A; \citealp{barbieri2020tweeteval}) & 166 & B & CC BY 3.0 (card) + Twitter ToS \\
spam\_\allowbreak{}phishing & Enron-Spam (\citealp{metsis2006spam}) & 167 & B & Not stated upstream (loader) \\
spam\_\allowbreak{}phishing & SMS Spam Collection (\citealp{almeida2011contributions}) & 167 & B & CC BY 4.0 (UCI) \\
spam\_\allowbreak{}phishing & SpamAssassin public corpus, ham and spam (2002-2005), and Jose Nazario's phishing corpus, 2004-2006 mailboxes & 166 & B & SpamAssassin corpus: not stated (read\dots{} \\
stance\_\allowbreak{}sarcasm\_\allowbreak{}humor & Humicroedit, SemEval-2020 Task 7 subtask 1 (\citealp{hossain2020semeval}) & 167 & A & Not stated (research data) (loader) \\
stance\_\allowbreak{}sarcasm\_\allowbreak{}humor & iSarcasmEval subtask A, English (\citealp{abufarha2022semeval}) & 167 & B & MIT \\
stance\_\allowbreak{}sarcasm\_\allowbreak{}humor & SemEval-2016 Task 6 stance detection (\citealp{mohammad2016semeval}), via TweetEval & 166 & B & Research use only (loader) \\
topic & 20 Newsgroups (\citealp{lang1995newsweeder}) & 167 & C & CC BY 4.0 (UCI) \\
topic & DBpedia hierarchical classes, level 2 (DeveloperOats/DBPedia\_Classes) & 167 & C & CC0 (card) \\
topic & Web of Science WOS-46985 (\citealp{kowsari2017hdltex}) & 166 & C & CC BY 4.0 \\
toxicity & Civil Comments / Jigsaw Unintended Bias (\citealp{borkan2019nuanced}) & 250 & A & CC0 \\
toxicity & Dynamically Generated Hate Speech Dataset v0.2.3 (\citealp{vidgen2021learning}) & 250 & A & CC BY 4.0 \\
\multicolumn{5}{l}{\textbf{Knowledge and reasoning}} \\
answerability & MuSiQue-Full, answerable and unanswerable multi-hop questions, validation (\citealp{trivedi2022musique}) & 250 & B & CC BY 4.0 \\
answerability & SQuAD 2.0, validation (\citealp{rajpurkar2018know}) & 250 & C & CC BY-SA 4.0 \\
causal & COPA, original test set with labels (\citealp{roemmele2011choice}), via Balanced COPA & 250 & B & CC BY 4.0 (card) \\
causal & e-CARE, validation (\citealp{du2022ecare}) & 250 & B & MIT \\
citation\_\allowbreak{}support & AttributionBench, in-distribution test (\citealp{li2024attributionbench}) & 250 & B & Apache-2.0 (component datasets keep t\dots{} \\
citation\_\allowbreak{}support & WiCE, Wikipedia citation entailment, sub-claims (\citealp{kamoi2023wice}) & 250 & B & ODC-BY (annotations) \\
commonsense & CommonsenseQA, validation (\citealp{talmor2019commonsenseqa}) & 167 & C & MIT \\
commonsense & PIQA, validation (\citealp{bisk2020piqa}) & 167 & C & AFL-3.0 \\
commonsense & WinoGrande XL, validation (\citealp{sakaguchi2020winogrande}) & 166 & C & CC BY \\
date\_\allowbreak{}reasoning & BIG-Bench Hard, date understanding (\citealp{suzgun2023challenging}) & 167 & C & MIT \\
date\_\allowbreak{}reasoning & Date arithmetic (generated) & 167 & C & CC BY 4.0 (generated by JevArena) \\
date\_\allowbreak{}reasoning & Timestamp comparison across time zones and formats (generated) & 166 & C & CC BY 4.0 (generated by JevArena) \\
fact\_\allowbreak{}verification & AVeriTeC real-world claims with question-answer evidence, dev (\citealp{schlichtkrull2023averitec}) & 167 & B & CC BY-NC 4.0 \\
fact\_\allowbreak{}verification & SciFact claims against paper abstracts, validation (\citealp{wadden2020fact}) & 167 & B & CC BY-NC 2.0 \\
fact\_\allowbreak{}verification & VitaminC contrastive fact verification (\citealp{schuster2021get}) & 166 & B & CC BY-SA 3.0 \\
factual\_\allowbreak{}judgment & Natural Questions open, validation (\citealp{lee2019latent}) & 250 & C & CC BY-SA 3.0 \\
factual\_\allowbreak{}judgment & SimpleQA Verified (Google, 2025; \citealp{wei2024measuring}) & 250 & C & MIT \\
logic\_\allowbreak{}puzzles & FOLIO v2, first-order logic reasoning in natural language, validation (\citealp{han2024folio}) & 167 & B & MIT \\
\bottomrule
\end{tabular}
\end{table*}

\begin{table*}[p]
\centering
\scriptsize
\setlength{\tabcolsep}{3pt}
\caption*{Table~\ref{tab:sources} (continued).}
\begin{tabular}{@{}>{\raggedright\arraybackslash}p{2.6cm}>{\raggedright\arraybackslash}p{7.0cm}rc>{\raggedright\arraybackslash}p{3.2cm}@{}}
\toprule
Task & Source & Requests & Tier & License \\
\midrule
logic\_\allowbreak{}puzzles & Ordering puzzles from pairwise clues (generated) & 167 & C & CC BY 4.0 (generated by JevArena) \\
logic\_\allowbreak{}puzzles & ZebraLogic, multiple-choice mode (\citealp{lin2025zebralogic}) & 166 & C & Not stated \\
math\_\allowbreak{}verification & ProcessBench model solutions (\citealp{zheng2025processbench}) & 250 & C & Apache-2.0 \\
math\_\allowbreak{}verification & VerifyBench, math problems (\citealp{yan2026verifybench}) & 250 & C & MIT \\
multihop & 2WikiMultihopQA, comparison and bridge-comparison questions (\citealp{ho2020constructing}) & 167 & C & Apache-2.0 \\
multihop & HotpotQA, distractor setting, comparison questions (\citealp{yang2018hotpotqa}) & 167 & C & CC BY-SA 4.0 \\
multihop & StrategyQA, re-split test (\citealp{geva2021aristotle}) & 166 & C & MIT \\
numeric\_\allowbreak{}comparison & EQUATE quantitative NLI: AWPNLI, NewsNLI, RedditNLI (\citealp{ravichander2019equate}) & 167 & B & MIT \\
numeric\_\allowbreak{}comparison & Number comparison across notations (generated) & 167 & C & CC BY 4.0 (generated by JevArena) \\
numeric\_\allowbreak{}comparison & Quantity comparison across units (generated) & 166 & C & CC BY 4.0 (generated by JevArena) \\
subject\_\allowbreak{}mcq & MMLU-Pro (\citealp{wang2024mmlupro}) & 250 & B & MIT \\
subject\_\allowbreak{}mcq & SuperGPQA (\citealp{map2025supergpqa}) & 250 & B & ODC-BY \\
\multicolumn{5}{l}{\textbf{Professional domains}} \\
answer\_\allowbreak{}grading & ELLIPSE corpus of English-learner essays, test set (\citealp{crossley2023english}) & 167 & A & CC BY-NC-SA 4.0 (loader) \\
answer\_\allowbreak{}grading & UNT Computer Science Short Answer Dataset v2.0 (\citealp{mohler2011learning}) & 167 & B & Not stated upstream (UNT release) \\
answer\_\allowbreak{}grading & SemEval-2013 Task 7 Student Response Analysis, SciEntsBank and Beetle, 5-way (\citealp{dzikovska2013semeval}) & 166 & A & CC BY-SA 3.0 \\
document\_\allowbreak{}fields & CORD v2 Consolidated Receipt Dataset, test and validation sets (Park et al., 2019) & 167 & B & CC BY 4.0 \\
document\_\allowbreak{}fields & SROIE, ICDAR 2019 Scanned Receipts OCR and Information Extraction, test set with the 14 restored annotations (\citealp{huang2019icdar2019}) & 167 & C & CC BY 4.0 \\
document\_\allowbreak{}fields & VRDU ad-buy forms (DeepForm political TV-ad invoices and contracts from FCC public files), mixed-template test split (\citealp{wang2023vrdu}) & 166 & B & Not stated (documents are US public r\dots{} \\
ecommerce & Shopping Queries Dataset (ESCI), English test split (\citealp{reddy2022shopping}) & 167 & A & Apache-2.0 \\
ecommerce & WANDS, Wayfair product search relevance judgements (\citealp{chen2022wands}) & 167 & A & MIT \\
ecommerce & WDC-25 Gold Standard for Product Categorization, test clusters (Web Data Commons, 2019) & 166 & C & Not stated (Web Data Commons \\
finance & Financial PhraseBank v1.0, sentences with majority agreement (\citealp{malo2014good}) & 167 & A & CC BY-NC-SA 3.0 \\
finance & Trillion Dollar Words: FOMC minutes, speeches and press conferences, hawkish-dovish test split (\citealp{shah2023trillion}) & 167 & A & CC BY-NC 4.0 \\
finance & South German Credit (UPDATE), 1,000 consumer credits 1973-1975 (Groemping, 2019; corrected Statlog German Credit) & 166 & B & CC BY 4.0 \\
legal & ContractNLI, document-level NLI on non-disclosure agreements, test split (\citealp{koreeda2021contractnli}) & 167 & B & CC BY 4.0 \\
legal & CUAD v1, clause-type identification on the 102 test contracts (\citealp{hendrycks2021cuad}) & 167 & B & CC BY 4.0 \\
legal & LegalBench, 17 classification tasks, test splits (\citealp{guha2023legalbench}) & 166 & B & Per task: CC BY 4.0 (most), CC BY-SA\dots{} \\
medical & Emergency department visits with expert-reviewed KTAS triage levels (Moon et al., 2019, PLOS ONE, S1 data) & 167 & B & CC BY 4.0 \\
medical & MedQA, US (USMLE) questions with 4 options, test split (\citealp{jin2021disease}) & 167 & C & MIT \\
medical & PubMedQA, expert-labelled PQA-L, official 500-question test set, reasoning-required setting (\citealp{jin2019pubmedqa}) & 166 & B & MIT \\
science & arXiv metadata, papers first submitted in 2024 with a computer-science primary category (Cornell arXiv snapshot) & 250 & A & CC0 1.0 (arXiv metadata, including ab\dots{} \\
science & ICLR 2025 submissions with reviewer ratings and decisions, from OpenReview via Paper Copilot paperlists & 250 & A & CC BY 4.0 (ICLR 2025 submissions and\dots{} \\
secops & NSL-KDD, KDDTest+ (\citealp{tavallaee2009detailed}; from the 1998 DARPA / KDD Cup 1999 data) & 167 & B & Redistribution permitted with citatio\dots{} \\
secops & NVD CVSS v3.1 base metrics assigned by NVD analysts, CVE-2025 records (NIST National Vulnerability Database) & 167 & A & CVE Terms of Use (descriptions,  MIT\dots{} \\
secops & NVD CWE mappings assigned by NVD analysts, CVE-2025 records (NIST National Vulnerability Database) & 166 & B & CVE Terms of Use (descriptions,  MIT\dots{} \\
support\_\allowbreak{}tickets & BANKING77, online banking customer queries, test split (\citealp{casanueva2020efficient}) & 167 & C & CC BY 4.0 \\
support\_\allowbreak{}tickets & CFPB Consumer Complaint Database, narratives of complaints received in 2022 (via a Hugging Face snapshot) & 167 & B & CC0 1.0 (US government public data) \\
support\_\allowbreak{}tickets & Mozilla Bugzilla defect reports filed January-June 2024 and later fixed (bugzilla.mozilla.org) & 166 & A & Not stated (public bug reports \\
\bottomrule
\end{tabular}
\end{table*}

\begin{table*}[p]
\centering
\scriptsize
\setlength{\tabcolsep}{3pt}
\caption*{Table~\ref{tab:sources} (continued).}
\begin{tabular}{@{}>{\raggedright\arraybackslash}p{2.6cm}>{\raggedright\arraybackslash}p{7.0cm}rc>{\raggedright\arraybackslash}p{3.2cm}@{}}
\toprule
Task & Source & Requests & Tier & License \\
\midrule
\multicolumn{5}{l}{\textbf{Retrieval, ranking and matching}} \\
context\_\allowbreak{}sufficiency & DuoRC ParaphraseRC, test (\citealp{saha2018duorc}) & 167 & C & MIT \\
context\_\allowbreak{}sufficiency & IIRC, incomplete information reading comprehension, test (\citealp{ferguson2020iirc}) & 166 & B & CC BY 4.0 \\
context\_\allowbreak{}sufficiency & MS MARCO question answering v2.1, dev set (\citealp{bajaj2018ms}) & 166 & A & MS MARCO terms: non-commercial resear\dots{} \\
entity\_\allowbreak{}matching & Abt-Buy (\citealp{kopcke2010evaluation}), DeepMatcher test split (\citealp{mudgal2018deep}) & 167 & B & Not stated (Leipzig entity-resolution\dots{} \\
entity\_\allowbreak{}matching & DBLP-Scholar (\citealp{kopcke2010evaluation}), DeepMatcher test split (\citealp{mudgal2018deep}) & 167 & B & Not stated (Leipzig entity-resolution\dots{} \\
entity\_\allowbreak{}matching & WDC Products, 80\% corner cases, seen-products test set (\citealp{peeters2024wdc}) & 166 & B & Not stated (offers extracted from the\dots{} \\
relevance\_\allowbreak{}grading & ANTIQUE non-factoid answer retrieval, test judgments (\citealp{hashemi2020antique}) & 167 & A & Research use (ANTIQUE readme) (loader) \\
relevance\_\allowbreak{}grading & TREC-COVID, final cumulative judgments (\citealp{voorhees2020treccovid}) & 167 & A & CORD-19 Dataset License Agreement (te\dots{} (loader) \\
relevance\_\allowbreak{}grading & TREC 2019 and 2020 Deep Learning Track, passage judgments on MS MARCO (\citealp{craswell2020overview}; 2021) & 166 & A & MS MARCO terms: non-commercial resear\dots{} \\
reranking & BRIGHT, StackExchange domains (\citealp{su2025bright}) & 250 & B & CC BY 4.0 \\
reranking & TREC 2019 and 2020 Deep Learning Track, passage judgments on MS MARCO (\citealp{craswell2020overview}; 2021) & 250 & B & MS MARCO terms: non-commercial resear\dots{} \\
retrieval\_\allowbreak{}need & FreshQA, test questions, snapshot of 2026-04-21 (\citealp{vu2024freshllms}) & 250 & A & Apache-2.0 \\
retrieval\_\allowbreak{}need & SituatedQA, temporal context-dependent question identification, test (\citealp{zhang2021situatedqa}) & 250 & B & Not stated (SituatedQA annotations) \\
\multicolumn{5}{l}{\textbf{Structured data, tables and logs}} \\
column\_\allowbreak{}types & SOTAB V2, WDC Schema.org Table Annotation Benchmark, column type annotation test set (Korini, Peeters and Bizer, 2023) & 167 & B & Not stated (Web Data Commons \\
column\_\allowbreak{}types & WDC Schema Matching Benchmark, SOTAB-SM (100-table variant), test split (Der and Bizer, 2023) & 167 & B & Not stated (Web Data Commons \\
column\_\allowbreak{}types & WDC Schema Matching Benchmark, T2D-SM with column headers, test split (Der and Bizer, 2023; \citealp{ritze2015matching}) & 166 & B & Not stated (Web Data Commons \\
data\_\allowbreak{}quality\_\allowbreak{}rules & Data-quality rules over generated tables (generated) & 250 & C & CC BY 4.0 (generated by JevArena) \\
data\_\allowbreak{}quality\_\allowbreak{}rules & HoloClean hospital data with injected typos and its denial constraints (\citealp{rekatsinas2017holoclean}) & 250 & B & Apache-2.0 (HoloClean repository) \\
log\_\allowbreak{}anomaly & Loghub BGL, Blue Gene/L RAS log with alert tags from the administrators' rules (\citealp{oliner2007supercomputers}) & 167 & B & CC BY 4.0 (Loghub, Zenodo record 8196\dots{} \\
log\_\allowbreak{}anomaly & Loghub BGL, alert categories assigned by the administrators' rules (\citealp{oliner2007supercomputers}) & 167 & B & CC BY 4.0 (Loghub, Zenodo record 8196\dots{} \\
log\_\allowbreak{}anomaly & Loghub HDFS\_v1, block-level traces labelled normal or anomalous (\citealp{xu2009detecting}) & 166 & C & CC BY 4.0 (Loghub, Zenodo record 8196\dots{} \\
sql\_\allowbreak{}verification & NL2SQL-BUGs, BIRD dev questions with SQL from RESDSQL, CodeS and CHESS (\citealp{liu2025nl2sqlbugs}) & 250 & B & Not stated (NL2SQL-BUGs labels) \\
sql\_\allowbreak{}verification & Spider dev with beam predictions of SmBoP, RESDSQL+NatSQL and NatSQL (\citealp{yu2018spider}; \citealp{chen2023error}) & 250 & B & Spider CC BY-SA 4.0 \\
table\_\allowbreak{}qa & TabMWP, multiple-choice questions of the test split (\citealp{lu2023dynamic}) & 167 & B & CC BY-NC-SA 4.0 \\
table\_\allowbreak{}qa & TAT-QA, test set with released gold answers (\citealp{zhu2021tatqa}) & 167 & C & CC BY 4.0 \\
table\_\allowbreak{}qa & WikiTableQuestions, pristine-unseen-tables test set (\citealp{pasupat2015compositional}) & 166 & C & CC BY-SA 4.0 \\
\multicolumn{5}{l}{\textbf{Code}} \\
code\_\allowbreak{}search & CodeSearchNet Challenge expert relevance annotations, six languages (\citealp{husain2019codesearchnet}) & 250 & A & MIT (annotations) \\
code\_\allowbreak{}search & CoSQA, Bing web queries paired with Python functions, dev split (\citealp{huang2021cosqa}) & 250 & A & MIT (CoCLR repository) \\
commit\_\allowbreak{}classification & CodeReviewer diff quality estimation, test split (\citealp{li2022automating}) & 250 & B & CC BY 4.0 (Zenodo record) \\
commit\_\allowbreak{}classification & CCS conventional commit classification dataset, test split (\citealp{zeng2025first}) & 250 & A & Apache-2.0 \\
fault\_\allowbreak{}localization & SWE-bench test set, file-level localization from the gold patch (\citealp{jimenez2024swebench}) & 167 & C & MIT (SWE-bench repository) \\
fault\_\allowbreak{}localization & SWE-bench test set, function-level localization inside the fixed file, oracle file contents (\citealp{jimenez2024swebench}) & 167 & C & MIT (SWE-bench repository) \\
fault\_\allowbreak{}localization & SWE-bench-Live, issues from 2024-2025, file-level localization from the gold patch (\citealp{zhang2025swebench}) & 166 & C & MIT \\
output\_\allowbreak{}prediction & CRUXEval-O, output prediction for 800 short Python functions (\citealp{gu2024cruxeval}) & 250 & B & MIT \\
output\_\allowbreak{}prediction & LiveCodeBench code execution v2, LeetCode solutions with recorded outputs (\citealp{jain2025livecodebench}) & 250 & B & Not stated precisely (the Hugging Fac\dots{} \\
\bottomrule
\end{tabular}
\end{table*}

\begin{table*}[p]
\centering
\scriptsize
\setlength{\tabcolsep}{3pt}
\caption*{Table~\ref{tab:sources} (continued).}
\begin{tabular}{@{}>{\raggedright\arraybackslash}p{2.6cm}>{\raggedright\arraybackslash}p{7.0cm}rc>{\raggedright\arraybackslash}p{3.2cm}@{}}
\toprule
Task & Source & Requests & Tier & License \\
\midrule
patch\_\allowbreak{}correctness & SWE-bench-extra tasks with SWE-agent final patches from Nebius's open-weight agent runs and their test results (Golubev et al., 2024) & 250 & C & CC BY 4.0 \\
patch\_\allowbreak{}correctness & SWE-rebench tasks with OpenHands + Qwen3-Coder-480B final patches and their test results (\citealp{badertdinov2025swerebench}) & 250 & C & CC BY 4.0 \\
static\_\allowbreak{}analysis\_\allowbreak{}triage & D2A leaderboard, Trace + Code task, dev split: Infer warnings on C projects labelled by differential analysis (\citealp{zheng2021d2a}) & 250 & B & Apache-2.0 (as listed by the Hugging\dots{} \\
static\_\allowbreak{}analysis\_\allowbreak{}triage & OWASP Benchmark for Java v1.2: 2,740 servlets, each a true or a false positive for one vulnerability category (OWASP Foundation) & 250 & C & GPL-2.0 \\
test\_\allowbreak{}outcome & HumanEvalFix from HumanEvalPack, correct and buggy solutions with their unit tests, six languages (\citealp{muennighoff2024octopack}) & 250 & C & MIT \\
test\_\allowbreak{}outcome & LiveCodeBench code generation, recorded per-test results of Gemini-Pro-1.5 programs (\citealp{jain2025livecodebench}) & 136 & C & MIT \\
vulnerability & PrimeVul v0.1, paired test set: vulnerable and patched C/C++ functions (\citealp{ding2025vulnerability}) & 250 & B & MIT \\
vulnerability & SecVulEval, C/C++ CVE fixes from NVD, vulnerable and fixed function pairs (\citealp{ahmed2026secvuleval}) & 250 & B & MIT \\
\multicolumn{5}{l}{\textbf{Forecasting}} \\
binary\_\allowbreak{}events & ForecastBench question sets, questions resolved from FRED, DBnomics, Wikipedia and Yahoo Finance data (\citealp{karger2025forecastbench}) & 167 & B & CC BY-SA 4.0 \\
binary\_\allowbreak{}events & ForecastBench question sets, prediction-market and forecasting-platform questions (\citealp{karger2025forecastbench}) & 167 & B & CC BY-SA 4.0 \\
binary\_\allowbreak{}events & Binary forecasting questions from Metaculus, Manifold, Polymarket, Good Judgment Open and INFER, test split (\citealp{halawi2024approaching}) & 166 & C & Apache-2.0 \\
multi\_\allowbreak{}outcome & Which related daily FRED series moves most (currencies against the dollar, energy prices, Treasury yields), resolved by September 2026 & 170 & C & Public-domain series from the Federal\dots{} \\
multi\_\allowbreak{}outcome & Polymarket multi-outcome events ended January-August 2026 (gamma API) & 250 & C & Polymarket Terms of Use (content rese\dots{} (loader) \\
numeric\_\allowbreak{}bins & Daily FRED series (Treasury yields, breakeven inflation, exchange rates, oil and gas prices), values published up to September 2026 & 250 & C & Public-domain series from the Federal\dots{} \\
numeric\_\allowbreak{}bins & Polymarket range ('bracket') events ended January-August 2026 (gamma API) & 250 & C & Polymarket Terms of Use (content rese\dots{} (loader) \\
\multicolumn{5}{l}{\textbf{Dialogue and interaction}} \\
clarification\_\allowbreak{}need & AmbigQA / AmbigNQ, validation (\citealp{min2020ambigqa}) & 250 & B & CC BY-SA 3.0 \\
clarification\_\allowbreak{}need & ShARC conversational machine reading with rules, test (\citealp{saeidi2018interpretation}) & 250 & B & CC BY-SA 3.0 \\
end\_\allowbreak{}of\_\allowbreak{}turn & Switchboard Dialog Act Corpus (SwDA), test conversations of Stolcke et al. (2000) (Jurafsky et al., 1997) & 250 & C & CC BY-NC-SA 3.0 \\
end\_\allowbreak{}of\_\allowbreak{}turn & Taskmaster-1 spoken two-person (Wizard-of-Oz) dialogues: user turns transcribed from speech (\citealp{byrne2019taskmaster1}) & 250 & C & CC BY 4.0 \\
human\_\allowbreak{}handoff & ABCD, Action-Based Conversations Dataset: customer-service chats escalated to a manager, all splits (\citealp{chen2021actionbased}) & 220 & B & MIT \\
human\_\allowbreak{}handoff & Scope cards over human intent labels: HINT3 v2 test sets, real queries to three chatbots (\citealp{arora2020hint3}), and NatCS customer-service requests in insurance, banking and finance (\citealp{gung2023natcs}) & 250 & C & HINT3: ODbL 1.0 (database) and DbCL 1\dots{} \\
state\_\allowbreak{}tracking & MultiWOZ 2.2, test split (\citealp{zang2020multiwoz}; \citealp{budzianowski2018multiwoz}) & 249 & B & MIT \\
state\_\allowbreak{}tracking & Schema-Guided Dialogue (SGD / DSTC8), test split (\citealp{rastogi2020scalable}) & 245 & B & CC BY-SA 4.0 \\
user\_\allowbreak{}satisfaction & EmoWOZ, test split: MultiWOZ and DialMAGE human-machine dialogues (\citealp{feng2022emowoz}) & 250 & A & CC BY-NC 4.0 \\
user\_\allowbreak{}satisfaction & User Satisfaction Simulation (USS), exchange-level ratings on MultiWOZ, SGD, ReDial and CCPE dialogues (\citealp{sun2021simulating}) & 250 & A & Not stated (source dialogues: MultiWO\dots{} \\
\multicolumn{5}{l}{\textbf{Batch semantic processing}} \\
semantic\_\allowbreak{}filter & Women's E-Commerce Clothing Reviews, 23,486 reviews from an online retailer (nicapotato, Kaggle) & 250 & C & CC0-1.0 \\
semantic\_\allowbreak{}filter & EMSCAD, Employment Scam Aegean Dataset: 17,880 job ads from Workable, 866 of them fraudulent (\citealp{vidros2017automatic}) & 250 & C & CC0-1.0 (Kaggle release) \\
semantic\_\allowbreak{}grouping & Amazon Berkeley Objects (ABO) product listings, English listings of amazon.com, .co.uk and .ca (\citealp{collins2022abo}) & 250 & C & CC BY 4.0 \\
semantic\_\allowbreak{}grouping & Twitter US Airline Sentiment, negative tweets with the complaint reason (Figure Eight / CrowdFlower Data for Everyone, 2015) & 250 & C & CC BY-NC-SA 4.0 \\
\bottomrule
\end{tabular}
\end{table*}

\begin{table*}[p]
\centering
\scriptsize
\setlength{\tabcolsep}{3pt}
\caption*{Table~\ref{tab:sources} (continued).}
\begin{tabular}{@{}>{\raggedright\arraybackslash}p{2.6cm}>{\raggedright\arraybackslash}p{7.0cm}rc>{\raggedright\arraybackslash}p{3.2cm}@{}}
\toprule
Task & Source & Requests & Tier & License \\
\midrule
semantic\_\allowbreak{}ranking & TREC 2022 Clinical Trials track: synthetic patient cases and ClinicalTrials.gov records with NIST judgments (\citealp{roberts2022overview}) & 250 & A & CC BY-SA 4.0 (Hugging Face copy) \\
semantic\_\allowbreak{}ranking & TripAdvisor hotel reviews with the reviewers' aspect ratings, English reviews of 2012 (\citealp{li2013identifying}; processed by Niimi, 2024) & 250 & B & Apache-2.0 (Hugging Face release) \\
\multicolumn{5}{l}{\textbf{LLM-as-judge and evaluators}} \\
agent\_\allowbreak{}trajectory & AgentRewardBench, expert-annotated web agent trajectories, test split (\citealp{lu2025agentrewardbench}) & 250 & A & AgentRewardBench terms of use: resear\dots{} \\
agent\_\allowbreak{}trajectory & tau-bench historical trajectories of GPT-4o and Claude 3.5 Sonnet (new) in the airline and retail domains (\citealp{yao2024taubench}) & 250 & C & MIT \\
atomic\_\allowbreak{}factuality & FActScore human-annotated biographies by InstructGPT, ChatGPT and PerplexityAI (\citealp{min2023factscore}) & 250 & B & MIT \\
atomic\_\allowbreak{}factuality & FELM, segment-level factuality of ChatGPT responses (\citealp{chen2023felm}) & 250 & B & CC BY-NC-SA 4.0 \\
instruction\_\allowbreak{}following & IFBench test prompts (\citealp{pyatkin2025generalizing}) with model responses and strict checker results from shisa-ai/eval-IFBench-results & 250 & B & ODC-BY (results) \\
instruction\_\allowbreak{}following & IFEval prompts (\citealp{zhou2023instructionfollowing}) with sampled responses and strict checker results from PPE (\citealp{frick2025evaluate}) & 250 & B & Apache-2.0 (prompts) \\
pairwise\_\allowbreak{}preference & JudgeBench, GPT-4o and Claude-3.5-Sonnet response pairs (\citealp{tan2025judgebench}) & 166 & A & MIT \\
pairwise\_\allowbreak{}preference & RewardBench 2, Math, Precise IF, Focus and Ties subsets (\citealp{malik2026rewardbench}) & 166 & B & ODC-BY \\
pairwise\_\allowbreak{}preference & RM-Bench, chat, math and code domains (\citealp{liu2025rmbench}) & 166 & B & ODC-BY \\
pointwise\_\allowbreak{}rating & BiGGen Bench, human evaluation split (\citealp{kim2025biggen}) & 250 & B & CC BY-SA 4.0 (BiGGen Bench) \\
pointwise\_\allowbreak{}rating & HelpSteer2, validation split with individual annotator ratings (\citealp{wang2024helpsteer}) & 250 & A & CC BY 4.0 \\
process\_\allowbreak{}reward & MR-Ben meta-reasoning benchmark, all subjects except coding (\citealp{zeng2024mrben}) & 167 & B & MIT \\
process\_\allowbreak{}reward & PRM800K, phase 2 test split (\citealp{lightman2024verify}) & 167 & B & MIT \\
process\_\allowbreak{}reward & ProcessBench, earliest-error paragraph (\citealp{zheng2025processbench}) & 166 & A & Apache-2.0 \\
rag\_\allowbreak{}faithfulness & BEGIN, knowledge-grounded dialogue responses, test split (\citealp{dziri2022evaluating}) & 250 & B & CC BY 4.0 (WoW and CMU-DoG parts) \\
rag\_\allowbreak{}faithfulness & RAGTruth, test split: QA (MS MARCO), data-to-text (Yelp) and news summarization (CNN/DM, Recent News) (\citealp{niu2024ragtruth}) & 250 & B & MIT \\
reference\_\allowbreak{}answer & Answer Equivalence (AE) dataset, test split, SQuAD questions (\citealp{bulian2022tomayto}) & 167 & A & Apache-2.0 (questions and passages fr\dots{} \\
reference\_\allowbreak{}answer & EVOUNA, Natural Questions and TriviaQA answers of five QA systems judged by humans (\citealp{wang2023evaluating}) & 167 & A & Apache-2.0 \\
reference\_\allowbreak{}answer & VerifierBench (CompassVerifier; \citealp{liu2025compassverifier}) & 166 & B & CC BY-SA 4.0 \\
summary\_\allowbreak{}consistency & FaithBench, summaries by 10 modern LLMs (\citealp{bao2025faithbench}) & 167 & B & CC BY-NC-SA 4.0 \\
summary\_\allowbreak{}consistency & SummEdits, 10 domains (\citealp{laban2023summedits}) & 167 & B & CC BY 4.0 \\
summary\_\allowbreak{}consistency & TofuEval, MeetingBank part, test split (\citealp{tang2024tofueval}) & 166 & A & MIT-0 (annotations) \\
translation\_\allowbreak{}quality & WMT 2020 newstest MQM annotations, en-de and zh-en (\citealp{freitag2021experts}) & 250 & B & Apache-2.0 (annotations) \\
translation\_\allowbreak{}quality & WMT 2023 general MT MQM annotations with three ratings per segment, en-de and zh-en (\citealp{kocmi2023findings}; Google MQM release) & 250 & B & Apache-2.0 (annotations) \\
\multicolumn{5}{l}{\textbf{Safety and moderation}} \\
agent\_\allowbreak{}action\_\allowbreak{}risk & ASSEBench safety and security subsets, unambiguous records (Luo et al., 2025, AgentAuditor) & 250 & B & Apache-2.0 \\
agent\_\allowbreak{}action\_\allowbreak{}risk & ATBench agent trajectory safety benchmark, 1,000-trajectory release (\citealp{li2026atbench}) & 250 & B & Apache-2.0 \\
custom\_\allowbreak{}policy & CoSApien, human-authored safety configs and test prompts (Zhang et al., ICLR 2025, Controllable Safety Alignment) & 167 & B & CDLA-Permissive-2.0 \\
custom\_\allowbreak{}policy & DynaBench test set of the DynaGuard project: business policies and user-agent dialogues (\citealp{hoover2026dynaguard}) & 167 & B & MIT \\
custom\_\allowbreak{}policy & Policy variants over the OpenAI moderation evaluation set (\citealp{markov2023holistic}; policies generated) & 166 & B & Not stated (the release repository ha\dots{} \\
harmfulness & Aegis 2.0 / Nemotron Content Safety Dataset V2, test, user prompts (\citealp{ghosh2025aegis20}) & 167 & B & CC BY 4.0 \\
harmfulness & BeaverTails-330k, test, question-answer pairs with three annotations each (\citealp{ji2023beavertails}) & 167 & A & CC BY-NC 4.0 \\
harmfulness & WildGuardTest, prompt and response harmfulness (\citealp{han2024wildguard}) & 166 & C & ODC-BY (AI2 Responsible Use Guidelines \\
\bottomrule
\end{tabular}
\end{table*}

\begin{table*}[p]
\centering
\scriptsize
\setlength{\tabcolsep}{3pt}
\caption*{Table~\ref{tab:sources} (continued).}
\begin{tabular}{@{}>{\raggedright\arraybackslash}p{2.6cm}>{\raggedright\arraybackslash}p{7.0cm}rc>{\raggedright\arraybackslash}p{3.2cm}@{}}
\toprule
Task & Source & Requests & Tier & License \\
\midrule
injection & AgentDojo tool-use trajectories of GPT-4o, clean and with injections in tool outputs (\citealp{debenedetti2024agentdojo}) & 166 & C & MIT \\
injection & BIPIA indirect prompt injection benchmark: email, table and code contexts with test attacks (\citealp{yi2025benchmarking}) & 166 & C & MIT (BIPIA) \\
injection & ToxicChat 0124, test, human-annotated user prompts with the jailbreaking label (\citealp{lin2023toxicchat}) & 166 & B & CC BY-NC 4.0 \\
pii & Text Anonymization Benchmark (TAB), test split: ECHR court judgments annotated for anonymization (\citealp{pilan2022text}) & 250 & B & MIT \\
pii & Customer messages, forms and log lines with inserted personal information (generated) & 250 & C & CC BY 4.0 (generated by JevArena) \\
refusal & CoCoNot, human-verified noncompliance test set and contrast set (\citealp{brahman2024art}) & 167 & B & ODC-BY (repository LICENSE.md) \\
refusal & XSTest v2 prompts: 250 safe prompts and 200 unsafe contrast prompts (\citealp{rottger2024xstest}) & 167 & B & CC BY 4.0 \\
refusal & XSTest v2 model completions with two human annotations each (\citealp{rottger2024xstest}) & 166 & A & CC BY 4.0 \\
\multicolumn{5}{l}{\textbf{Routing and orchestration}} \\
context\_\allowbreak{}pruning & Leave-one-out paragraph ablation on generated multi-hop fact chains (generated) & 250 & C & CC BY 4.0 (generated by JevArena) \\
context\_\allowbreak{}pruning & QASPER, questions on NLP papers with annotated evidence paragraphs, test split (\citealp{dasigi2021dataset}) & 250 & B & CC BY 4.0 \\
difficulty & BigCodeBench-Instruct tasks with leaderboard solve rates (\citealp{zhuo2025bigcodebench}) & 167 & C & Apache-2.0 \\
difficulty & EmbedLLM correctness matrix, test questions (\citealp{zhuang2025embedllm}) & 167 & C & Apache-2.0 (card) \\
difficulty & LiveBench code generation, code completion, typo fixing and Connections puzzles (\citealp{white2025livebench}) & 166 & C & Not stated (dataset cards) \\
escalation & AutoMix cascade outputs on QuALITY, validation (\citealp{aggarwal2024automix}) & 167 & B & Apache-2.0 (AutoMix) \\
escalation & LiveBench model answers and judgments: GPT-4o mini drafts, o1-preview as the stronger model (\citealp{white2025livebench}) & 167 & C & Not stated (dataset cards) \\
escalation & RouterArena cached GPT-4o mini replies, full split (\citealp{lu2026routerarena}) & 166 & C & Apache-2.0 (RouterArena repository) \\
memory\_\allowbreak{}decisions & DuLeMon, Chinese long-term persona memory dialogues, 'both' test split (\citealp{xu2022long}) & 250 & C & Not stated (data) \\
memory\_\allowbreak{}decisions & Multi-Session Chat, per-turn memory summaries, test split (\citealp{xu2022beyond}) & 250 & B & Not stated (data) \\
model\_\allowbreak{}correctness & LiveBench code generation, code completion, typo fixing and Connections puzzles (\citealp{white2025livebench}) & 167 & C & Not stated (dataset cards) \\
model\_\allowbreak{}correctness & RouterArena sub\_10 split with cached replies of three models (\citealp{lu2026routerarena}) & 167 & C & Apache-2.0 (RouterArena repository) \\
model\_\allowbreak{}correctness & RouterBench, zero-shot MMLU, HellaSwag and ARC-Challenge prompts (\citealp{hu2024routerbench}) & 166 & C & Not stated (dataset card) \\
model\_\allowbreak{}selection & LiveBench code generation, code completion, typo fixing and Connections puzzles (\citealp{white2025livebench}) & 212 & B & Not stated (dataset cards) \\
model\_\allowbreak{}selection & RouterBench, zero-shot MMLU, HellaSwag and ARC-Challenge prompts (\citealp{hu2024routerbench}) & 250 & B & Not stated (dataset card) \\
tool\_\allowbreak{}selection & Berkeley Function Calling Leaderboard v4, expert-written 'multiple' and 'irrelevance' categories (\citealp{patil2025berkeley}) & 167 & B & Apache-2.0 \\
tool\_\allowbreak{}selection & Berkeley Function Calling Leaderboard v4, user-contributed 'live\_multiple' and 'live\_irrelevance' categories (\citealp{patil2025berkeley}) & 167 & B & Apache-2.0 \\
tool\_\allowbreak{}selection & MetaTool ToolE, tool selection with and without the right tool in the list (\citealp{huang2024metatool}) & 166 & A & MIT \\
\multicolumn{5}{l}{\textbf{Subjective and pluralistic}} \\
moral\_\allowbreak{}judgment & ETHICS commonsense morality, hard test set, short scenarios (\citealp{hendrycks2021aligning}) & 167 & B & MIT \\
moral\_\allowbreak{}judgment & Scruples Anecdotes, test split: r/AmItheAsshole posts with the community's verdict votes (\citealp{lourie2021scruples}) & 167 & B & Apache-2.0 \\
moral\_\allowbreak{}judgment & Social-Chem-101, rules of thumb annotated by 50 crowd workers each (analysis split; \citealp{forbes2020social}) & 166 & A & CC BY-SA 4.0 \\
opinion\_\allowbreak{}distribution & Collective Constitutional AI: Polis votes of about 1,000 U.S. adults on rules for an AI chatbot, July 2023 (Anthropic and The Collective Intelligence Project, 2023) & 125 & B & Not stated \\
opinion\_\allowbreak{}distribution & GlobalOpinionQA: Pew Global Attitudes Survey and World Values Survey questions with answer shares by country (\citealp{durmus2023measuring}) & 125 & B & CC BY-NC-SA 4.0 \\
opinion\_\allowbreak{}distribution & Polis open data: statements from 13 English-language public deliberations with participants' agree / disagree / pass votes (The Computational Democracy Project) & 125 & B & CC BY 4.0 \\
opinion\_\allowbreak{}distribution & SubPOP-Eval: General Social Survey 2022 questions with answer shares of 22 U.S. subpopulations (\citealp{suh2025language}) & 125 & B & CC BY-NC-SA 4.0 (gated (loader) \\
\bottomrule
\end{tabular}
\end{table*}

\begin{table*}[p]
\centering
\scriptsize
\setlength{\tabcolsep}{3pt}
\caption*{Table~\ref{tab:sources} (continued).}
\begin{tabular}{@{}>{\raggedright\arraybackslash}p{2.6cm}>{\raggedright\arraybackslash}p{7.0cm}rc>{\raggedright\arraybackslash}p{3.2cm}@{}}
\toprule
Task & Source & Requests & Tier & License \\
\midrule
perspective\_\allowbreak{}labeling & DICES-350: adversarial chatbot conversations rated for safety by all 104 raters of a pool balanced by gender and race (\citealp{aroyo2023dices}) & 167 & B & CC BY 4.0 \\
perspective\_\allowbreak{}labeling & DICES-990: adversarial chatbot conversations, each rated for safety by 60-70 raters balanced by gender and locale (US, India) (\citealp{aroyo2023dices}) & 167 & B & CC BY 4.0 \\
perspective\_\allowbreak{}labeling & Measuring Hate Speech corpus, the 70 reference-set comments rated by 240-815 U.S. crowd workers each (\citealp{kennedy2020measuring}; \citealp{sachdeva2022measuring}) & 166 & A & CC BY 4.0 \\
persuasion\_\allowbreak{}humor & Persuasion dataset: claims with human- and Claude-written arguments, each read by three participants (Durmus et al., 2024) & 125 & B & CC BY-NC-SA 4.0 \\
persuasion\_\allowbreak{}humor & ExPUN: puns and non-puns from SemEval-2017 Task 7, each rated by five annotators (\citealp{sun2022expunations}) & 125 & B & CC BY-NC 4.0 (annotations) \\
persuasion\_\allowbreak{}humor & New Yorker Cartoon Caption Contest crowd ratings (Jain et al., 2020; \citealp{zhang2024humor}), cartoons described by the human-written annotations of Hessel et al. (2023), ranking test split & 125 & B & CC BY-NC 4.0 (ratings) \\
persuasion\_\allowbreak{}humor & UKPConvArg1, argument pairs from createdebate.com and convinceme.net with five crowd judgments each (\citealp{habernal2016argument}) & 124 & A & CC BY 4.0 \\
\multicolumn{5}{l}{\textbf{Agent single steps}} \\
action\_\allowbreak{}critic & AgentProcessBench, HotpotQA and GAIA (text) subsets, human step labels (2026) & 167 & B & MIT \\
action\_\allowbreak{}critic & TRBench-BFCL, single-turn parallel and parallel-multiple categories (Li et al., 2025, ToolRM) & 167 & B & Apache-2.0 (Hugging Face card and LIC\dots{} \\
action\_\allowbreak{}critic & WebRewardBench, Mind2Web test steps with expert and model-proposed actions (Chae et al., 2025, Web-Shepherd) & 166 & C & Not stated (WebRewardBench) (loader) \\
completion & tau2-bench telecom domain in solo mode (no user), published simulations of GPT-4.1 and o4-mini agents (\citealp{barres2025tau2}) & 250 & C & MIT \\
completion & Terminal-Bench 2.0 leaderboard runs of the Terminus-2 agent with eight models (\citealp{merrill2026terminalbench}) & 250 & C & Apache-2.0 \\
failure\_\allowbreak{}attribution & AgentErrorBench, failed ALFWorld, WebShop and GAIA runs with expert root-cause annotation (Zhu et al., 2025, AgentDebug) & 167 & B & MIT (repository license \\
failure\_\allowbreak{}attribution & TRAIL, expert-annotated error taxonomy over GAIA and SWE-bench agent traces (\citealp{deshpande2025trail}) & 148 & B & MIT, gated: users agree not to reshar\dots{} (loader) \\
failure\_\allowbreak{}attribution & Who\&When, failed multi-agent logs on GAIA and AssistantBench with annotated failure agent and step (\citealp{zhang2025agent}) & 166 & B & MIT (GitHub repository ag2ai/Agents\_F\dots{} \\
next\_\allowbreak{}action & AndroidControl official test split, human demonstrations on 833 Android apps (\citealp{li2024effects}) & 250 & B & Apache-2.0 (google-research repositor\dots{} \\
next\_\allowbreak{}action & Mind2Web test splits (cross-task, cross-website, cross-domain) in the human-verified Multimodal-Mind2Web release (\citealp{deng2023mind2web}) & 250 & B & CC BY 4.0 (Mind2Web (loader) \\
progress\_\allowbreak{}success & SWE-rebench July 2026 leaderboard trajectories: 111 fresh issues, agentic systems and models in a common scaffold (Nebius, 2026) & 167 & C & Other: the release does not relicense\dots{} \\
progress\_\allowbreak{}success & tau2-bench telecom domain, published simulations of four agent models with a GPT-4.1 user simulator (\citealp{barres2025tau2}) & 167 & C & MIT \\
progress\_\allowbreak{}success & Terminal-Bench 2.0 leaderboard runs of the Terminus-2 agent with eight models (\citealp{merrill2026terminalbench}) & 166 & C & Apache-2.0 \\
step\_\allowbreak{}correctness & ToolComp process-supervision pairs, human-corrected versus original ReAct steps and action plans (\citealp{nath2025toolcomp}) & 250 & B & Not stated (the sibling repository va\dots{} \\
step\_\allowbreak{}correctness & TRBench-BFCL, multi-turn base category (Li et al., 2025, ToolRM), built on BFCL v3 multi-turn & 250 & B & Apache-2.0 (Hugging Face card and LIC\dots{} \\
tool\_\allowbreak{}arguments & Berkeley Function Calling Leaderboard v4, user-contributed 'live\_simple' category (\citealp{patil2025berkeley}) & 167 & B & Apache-2.0 \\
tool\_\allowbreak{}arguments & Berkeley Function Calling Leaderboard v4, expert-written 'simple' categories in Python, Java and JavaScript (\citealp{patil2025berkeley}) & 167 & B & Apache-2.0 \\
tool\_\allowbreak{}arguments & ComplexFuncBench, multi-step Booking.com function calling with recorded API results (\citealp{zhong2025complexfuncbench}) & 166 & C & Apache-2.0 \\
\multicolumn{5}{l}{\textbf{Games (static)}} \\
chess\_\allowbreak{}best\_\allowbreak{}move & ChessBench test positions with Stockfish 16 action values (\citealp{ruoss2024amortized}) & 250 & B & CC BY 4.0 (ChessBench annotations) \\
chess\_\allowbreak{}best\_\allowbreak{}move & Lichess puzzle database, first shard of the September 2026 export (lichess.org) & 250 & A & CC0 1.0 \\
poker & PokerBench post-flop test set, solutions from WASM-Postflop (\citealp{zhuang2025pokerbench}) & 250 & A & Apache-2.0 \\
poker & PokerBench pre-flop test set, solutions from GTO Wizard charts (\citealp{zhuang2025pokerbench}) & 250 & A & Apache-2.0 \\
\bottomrule
\end{tabular}
\end{table*}

\begin{table*}[p]
\centering
\scriptsize
\setlength{\tabcolsep}{3pt}
\caption*{Table~\ref{tab:sources} (continued).}
\begin{tabular}{@{}>{\raggedright\arraybackslash}p{2.6cm}>{\raggedright\arraybackslash}p{7.0cm}rc>{\raggedright\arraybackslash}p{3.2cm}@{}}
\toprule
Task & Source & Requests & Tier & License \\
\midrule
position\_\allowbreak{}eval & ChessBench test positions with Stockfish 16 state values (\citealp{ruoss2024amortized}) & 167 & B & CC BY 4.0 (ChessBench annotations) \\
position\_\allowbreak{}eval & Connect-4 opening database: all 8-ply positions with their game-theoretic value (John Tromp, UCI repository) & 167 & C & CC BY 4.0 \\
position\_\allowbreak{}eval & Lichess rated standard games of 1 June 2024, first shard of the June 2024 export (lichess.org) & 166 & C & CC0 1.0 \\
puzzles & Minesweeper Beginner positions (9x9, 10 mines) with an exact deduction per square (generated) & 167 & C & CC BY 4.0 (generated by JevArena) \\
puzzles & 8-puzzle positions with a unique best first move (generated) & 167 & C & CC BY 4.0 (generated by JevArena) \\
puzzles & Sudoku grids: rule check and completability (generated) & 166 & C & CC BY 4.0 (generated by JevArena) \\
social\_\allowbreak{}games & It Takes Two to Lie: Diplomacy negotiation messages with the senders' own truthfulness labels, test games (\citealp{peskov2020takes}) & 167 & B & CC BY 4.0 \\
social\_\allowbreak{}games & AH2AC2 human Hanabi games from hanab.live, 2- and 3-player (\citealp{dizdarevic2025adhoc}), with exact card-counting beliefs & 167 & C & Apache-2.0 \\
social\_\allowbreak{}games & FanLang-9: human 9-player Werewolf games with roles, votes and transcribed speeches (\citealp{wu2024enhance}) & 166 & C & MIT \\
\multicolumn{5}{l}{\textbf{Embodied and robotics (static)}} \\
action\_\allowbreak{}safety & ASIMOV 2.0 injury scenarios: effect of a robot's instruction on the chance of injury (\citealp{jindal2025ai}) & 167 & B & Not stated \\
action\_\allowbreak{}safety & Robot drive commands on a room floor plan (generated) & 167 & C & CC BY 4.0 (generated by JevArena) \\
action\_\allowbreak{}safety & SafeText: situations with safe and physically unsafe advice (\citealp{levy2022safetext}) & 166 & B & MIT \\
failure\_\allowbreak{}type & ACPBench, test and validation splits (\citealp{kokel2025acpbench}): plan validation, multiple-choice form (failure category) and the first inapplicable action (failure step) & 250 & B & CDLA-Permissive-2.0 \\
failure\_\allowbreak{}type & PlanBench plan verification task, Blocksworld and Logistics (\citealp{valmeekam2023planbench}) & 250 & B & MIT \\
feasibility & ACPBench, test and validation splits (\citealp{kokel2025acpbench}): action applicability, yes/no and multiple-choice forms & 167 & B & CDLA-Permissive-2.0 \\
feasibility & AI2-THOR object metadata (ProcTHOR's dump of 8,715 objects in 195 scenes) and placement restrictions (SimObjType.cs) & 167 & C & Apache-2.0 \\
feasibility & VirtualHome object properties (properties\_data.json) and the action checks of its evolving-graph simulator (\citealp{puig2018virtualhome}) & 166 & C & MIT \\
next\_\allowbreak{}skill & ACPBench Hard, next action towards the goal, test and validation splits (\citealp{kokel2026acpbench}) & 159 & B & CDLA-Permissive-2.0 \\
next\_\allowbreak{}skill & ALFWorld text games built on ALFRED scenes, PDDL problems of the 0.4.2 release (\citealp{shridhar2021alfworld}) & 167 & C & MIT \\
next\_\allowbreak{}skill & PlanBench Blocksworld problems, next action on a shortest plan (\citealp{valmeekam2023planbench}; answers by exhaustive search) & 166 & C & MIT \\
spatial\_\allowbreak{}reasoning & LLM-BabyBench Predict: final agent pose after an action sequence in BabyAI grid worlds (2025) & 167 & B & MIT \\
spatial\_\allowbreak{}reasoning & SpatialEval text-only split, Spatial-Map and Maze-Nav tasks (\citealp{wang2024picture}) & 167 & A & Not stated \\
spatial\_\allowbreak{}reasoning & StepGame, test set with distracting statements (\citealp{shi2022stepgame}; regenerated February 2024) & 166 & B & MIT \\
task\_\allowbreak{}success & ACPBench, test and validation splits (\citealp{kokel2025acpbench}): plan validation, yes/no form & 167 & B & CDLA-Permissive-2.0 \\
task\_\allowbreak{}success & EB-ALFRED trajectory dataset of EmbodiedBench: 11 vision-language models in AI2-THOR (\citealp{yang2025embodiedbench}) & 167 & B & Not stated \\
task\_\allowbreak{}success & PlanBench Blocksworld plan generation: plans written by 13 language models with VAL checks (Valmeekam et al., 2023-2025) & 166 & B & MIT \\
\bottomrule
\end{tabular}
\end{table*}

\section{Closed-Loop Environments}
\label{app:envs}

Our environments come from agent benchmarks for the web (MiniWoB++ \citep{shi2017wob,liu2018miniwob}, WebShop
\citep{yao2022webshop}), text games (TextWorld \citep{cote2018textworld}, Jericho \citep{hausknecht2020jericho}, ScienceWorld
\citep{wang2022scienceworld}, ALFWorld \citep{shridhar2021alfworld}), support agents ($\tau^2$-bench \citep{barres2025tau2}),
coding agents (SWE-bench \citep{jimenez2024swebench}), the long-horizon games of BALROG \citep{paglieri2025balrog}, and
control and game simulators built for reinforcement learning (MuJoCo \citep{towers2024gymnasium,todorov2012mujoco},
Meta-World \citep{yu2019metaworld}, the Arcade Learning Environment \citep{bellemare2013ale}, ViZDoom
\citep{kempka2016vizdoom}, highway-env \citep{leurent2018highway}, and PyFlyt \citep{tai2023pyflyt}).
Table~\ref{tab:envs} lists every environment. A failed answer falls back to a fixed default action. Outcomes are the
simulator's own: success where the task defines it, and otherwise its return, score or progression.

\paragraph{Coding controller.} The coding controller turns SWE-bench Verified \citep{jimenez2024swebench,openai2024sweverified}
into decisions. A fixed proposer inside the environment writes the candidate next actions (code views, test runs, patches, a
revert); the model picks one, which runs in the task's official container, or submits the changes so far, which SWE-bench's
own tests grade. The scripted proposer offers the reference patch among failing patches from public submissions, so the task
is to recognize the right fix and stop; the GPT-5.6 Luna proposer writes five actions without reasoning and never the
reference patch, so the decisions have to steer it toward a fix. This proposer is the only LLM inside a harness.

\paragraph{Realtime mode.} Each model $m$ is charged a fixed latency $L_m$ per decision: its median latency per request,
measured on an idle A100 GPU and rounded to 10~ms. While a decision is pending, the environment advances
$\operatorname{round}(L_m / \Delta t)$ steps of length $\Delta t$ with the previous action held. A fixed latency, rather than
the measured wall-clock time, keeps episodes reproducible. A two-car intersection extends this mode to several agents, each
controlled by an independent call of the same model.

\paragraph{Action encodings.} The action encodings follow the simulators. In MuJoCo each actuator is a 5-level score question from the strongest
negative to the strongest positive torque; in Meta-World the model picks a motion primitive that a scripted controller executes;
in MiniWoB++ it picks an operation, a target element and a text; in $\tau^2$-bench it first picks one of 43 tools or ``done''
and then every argument from candidates taken from the tool schema, the ticket, the policy and earlier results; in chess it picks
one of the legal moves against Stockfish \citep{stockfish}; in text games and Wikispeedia \citep{west2009wikispeedia} it picks a
command or a link, where the commands of Jericho's text adventures \citep{hausknecht2020jericho} are those its valid-action
detector finds, plus ``look'' and ``wait''. In WebShop \citep{yao2022webshop} the model picks one of the page's clickable actions,
or on the search page one of two to four queries derived from the instruction by fixed rules. In ScienceWorld
\citep{wang2022scienceworld} an action takes up to three questions, first its kind and then its objects, each built from the
simulator's list of valid actions. The games from BALROG \citep{paglieri2025balrog} (BabyAI, Crafter \citep{hafner2022crafter},
Baba Is AI \citep{cloos2024baba}, MiniHack \citep{samvelyan2021minihack} and NetHack \citep{kuttler2020nle}) keep BALROG's text
observations, and every key press is one choice from BALROG's action list.

Configuration changes relative to the original simulators are limited to what
the decision interface needs: the quadrotor flies in PyFlyt's position mode with a sparse reward and a 30-second limit, the
two-car intersection uses destinations whose routes cross or share an exit, and the $\tau^2$ telecom domain runs in its no-user
mode with two small fixes to the environment's construction. Atari games report the ALE game score, since they define no success.

The newer suites keep their benchmarks' settings where possible. WebShop plays the first 100 of its 500 test goals through its
own text environment; the search page offers two to four queries derived from the instruction. ScienceWorld plays one test
variation per task type with its ``easy'' simplifications and a limit of 100 actions. The BALROG games keep BALROG's
observations, action lists and step limits, with fixes that make episodes reproducible: Crafter keeps its creatures in insertion
order rather than in memory-address order, and NetHack runs with a pinned clock, since the game reads the date (moon phase,
Friday the 13th). Jericho offers the commands its valid-action detector finds, plus ``look'' and ``wait'', for 100 commands per
game. The coding controller plays 40 SWE-bench Verified instances, balanced across repositories and validated before use, in
their official evaluation images, and masks timings, process ids, addresses and temporary paths in command outputs so that the
same actions give the same text.

\begin{table*}[p]
\centering
\scriptsize
\setlength{\tabcolsep}{3pt}
\caption{\textbf{All closed-loop environments.} Seeds are 0 to episodes$-$1; ``Jev'' is success rate / mean score over all seeds (``--'' where success is undefined or runs are pending).}
\label{tab:envs}
\begin{tabular}{@{}>{\raggedright\arraybackslash}p{3.4cm}>{\raggedright\arraybackslash}p{5.4cm}lrr>{\raggedright\arraybackslash}p{1.5cm}r@{}}
\toprule
Environment & Source & Mode & Episodes & Max dec. & Score & Jev \\
\midrule
\multicolumn{7}{@{}l}{\textbf{embodied/household\_loop}} \\
alfworld & ALFWorld (\citealp{shridhar2021alfworld}), TextWorld version, out-of-distribution evaluation games (valid\_unseen) & paused & 24 & 50 & won (1) or not (0) & 0.46 / 0.5 \\
\multicolumn{7}{@{}l}{\textbf{embodied/mujoco\_control}} \\
half\_\allowbreak{}cheetah & Gymnasium HalfCheetah-v5 (MuJoCo) & paused & 10 & 300 & return & -- / -44.9 \\
hopper & Gymnasium Hopper-v5 (MuJoCo) & paused & 10 & 250 & return & -- / 306.6 \\
inverted\_\allowbreak{}double\_\allowbreak{}pendulum & Gymnasium InvertedDoublePendulum-v5 (MuJoCo) & paused & 10 & 500 & return & 0.00 / 75.2 \\
inverted\_\allowbreak{}pendulum & Gymnasium InvertedPendulum-v5 (MuJoCo) & paused & 10 & 500 & return & 0.00 / 20.7 \\
reacher & Gymnasium Reacher-v5 (MuJoCo) & paused & 10 & 50 & return & -- / -34.5 \\
\multicolumn{7}{@{}l}{\textbf{embodied/tabletop\_loop}} \\
button\_\allowbreak{}press\_\allowbreak{}topdown & Meta-World button-press-topdown-v3 (MuJoCo) & paused & 10 & 80 & success & 0.40 / 0.4 \\
door\_\allowbreak{}open & Meta-World door-open-v3 (MuJoCo) & paused & 10 & 80 & success & 0.10 / 0.1 \\
drawer\_\allowbreak{}close & Meta-World drawer-close-v3 (MuJoCo) & paused & 10 & 40 & success & 1.00 / 1.0 \\
drawer\_\allowbreak{}open & Meta-World drawer-open-v3 (MuJoCo) & paused & 10 & 80 & success & 0.00 / 0.0 \\
peg\_\allowbreak{}insert\_\allowbreak{}side & Meta-World peg-insert-side-v3 (MuJoCo) & paused & 10 & 80 & success & 0.00 / 0.0 \\
pick\_\allowbreak{}place & Meta-World pick-place-v3 (MuJoCo) & paused & 10 & 80 & success & 0.00 / 0.0 \\
push & Meta-World push-v3 (MuJoCo) & paused & 10 & 80 & success & 0.00 / 0.0 \\
reach & Meta-World reach-v3 (MuJoCo) & paused & 10 & 80 & success & 1.00 / 1.0 \\
window\_\allowbreak{}close & Meta-World window-close-v3 (MuJoCo) & paused & 10 & 80 & success & 1.00 / 1.0 \\
window\_\allowbreak{}open & Meta-World window-open-v3 (MuJoCo) & paused & 10 & 80 & success & 1.00 / 1.0 \\
\multicolumn{7}{@{}l}{\textbf{game/chess\_play}} \\
stockfish\_\allowbreak{}level0 & Stockfish 19 at skill level 0 (the weakest official setting), move choice re-implemented with a seeded generator & paused & 20 & 150 & result & 0.00 / 0.0 \\
\multicolumn{7}{@{}l}{\textbf{game/open\_world}} \\
crafter & Crafter (\citealp{hafner2022crafter}; crafter 1.8.3), default world, through BALROG's text wrapper & paused & 10 & 2000 & progression (\% of 22 achievements) & -- / 5.5 \\
\multicolumn{7}{@{}l}{\textbf{game/roguelike}} \\
babyai\_\allowbreak{}goto\_\allowbreak{}local & BabyAI GoToLocal (BabyAI-GoToLocal-v0, Minigrid) & paused & 10 & 64 & reward & 0.30 / 0.3 \\
babyai\_\allowbreak{}goto\_\allowbreak{}obj & BabyAI GoToObj (BabyAI-GoToObj-v0, Minigrid) & paused & 10 & 64 & reward & 0.70 / 0.6 \\
babyai\_\allowbreak{}mini\_\allowbreak{}boss\_\allowbreak{}level & BabyAI MiniBossLevel (BabyAI-MiniBossLevel-v0, Minigrid) & paused & 10 & 800 & reward & 0.00 / 0.0 \\
babyai\_\allowbreak{}pickup\_\allowbreak{}loc & BabyAI PickupLoc (BabyAI-PickupLoc-v0, Minigrid) & paused & 10 & 64 & reward & 0.20 / 0.2 \\
babyai\_\allowbreak{}put\_\allowbreak{}next\_\allowbreak{}local & BabyAI PutNextLocal (BabyAI-PutNextLocal-v0, Minigrid) & paused & 10 & 128 & reward & 0.00 / 0.0 \\
babyai\_\allowbreak{}unlock\_\allowbreak{}pickup & BabyAI UnlockPickup (BabyAI-UnlockPickup-v0, Minigrid) & paused & 10 & 72 & reward & 0.00 / 0.0 \\
minihack\_\allowbreak{}boxoban\_\allowbreak{}hard & MiniHack MiniHack-Boxoban-Hard-v0 (\citealp{samvelyan2021minihack}; BALROG's fork 3ecb6da on balrog-nle 0.9.0), a BALROG task & paused & 10 & 100 & progression (\%: 100 if solved) & 0.00 / 0.0 \\
minihack\_\allowbreak{}boxoban\_\allowbreak{}medium & MiniHack MiniHack-Boxoban-Medium-v0 (\citealp{samvelyan2021minihack}; BALROG's fork 3ecb6da on balrog-nle 0.9.0), a BALROG task & paused & 10 & 100 & progression (\%: 100 if solved) & 0.00 / 0.0 \\
minihack\_\allowbreak{}corridor\_\allowbreak{}r3 & MiniHack MiniHack-Corridor-R3-v0 (\citealp{samvelyan2021minihack}; BALROG's fork 3ecb6da on balrog-nle 0.9.0), a BALROG task & paused & 10 & 100 & progression (\%: 100 if solved) & 0.00 / 0.0 \\
minihack\_\allowbreak{}corridorbattle\_\allowbreak{}dark & MiniHack MiniHack-CorridorBattle-Dark-v0 (\citealp{samvelyan2021minihack}; BALROG's fork 3ecb6da on balrog-nle 0.9.0), a BALROG task & paused & 10 & 100 & progression (\%: 100 if solved) & 0.00 / 0.0 \\
minihack\_\allowbreak{}mazewalk\_\allowbreak{}15x15 & MiniHack MiniHack-MazeWalk-15x15-v0 (\citealp{samvelyan2021minihack}; BALROG's fork 3ecb6da on balrog-nle 0.9.0), a BALROG task & paused & 10 & 100 & progression (\%: 100 if solved) & 0.00 / 0.0 \\
minihack\_\allowbreak{}mazewalk\_\allowbreak{}9x9 & MiniHack MiniHack-MazeWalk-9x9-v0 (\citealp{samvelyan2021minihack}; BALROG's fork 3ecb6da on balrog-nle 0.9.0), a BALROG task & paused & 10 & 100 & progression (\%: 100 if solved) & 0.40 / 40.0 \\
minihack\_\allowbreak{}quest\_\allowbreak{}easy & MiniHack MiniHack-Quest-Easy-v0 (\citealp{samvelyan2021minihack}; BALROG's fork 3ecb6da on balrog-nle 0.9.0), a BALROG task & paused & 10 & 100 & progression (\%: 100 if solved) & 0.00 / 0.0 \\
minihack\_\allowbreak{}quest\_\allowbreak{}medium & MiniHack MiniHack-Quest-Medium-v0 (\citealp{samvelyan2021minihack}; BALROG's fork 3ecb6da on balrog-nle 0.9.0), a BALROG task & paused & 10 & 100 & progression (\%: 100 if solved) & 0.00 / 0.0 \\
nethack & NetHack 3.6.6 in the NetHack Learning Environment, NetHackChallenge-v0 (\citealp{kuttler2020nle}; balrog-nle 0.9.0), a BALROG task & paused & 5 & 2000 & progression (\%, BALROG's NetHack progression) & -- / 0.0 \\
\bottomrule
\end{tabular}
\end{table*}

\begin{table*}[p]
\centering
\scriptsize
\setlength{\tabcolsep}{3pt}
\caption*{Table~\ref{tab:envs} (continued).}
\begin{tabular}{@{}>{\raggedright\arraybackslash}p{3.4cm}>{\raggedright\arraybackslash}p{5.4cm}lrr>{\raggedright\arraybackslash}p{1.5cm}r@{}}
\toprule
Environment & Source & Mode & Episodes & Max dec. & Score & Jev \\
\midrule
\multicolumn{7}{@{}l}{\textbf{game/rule\_puzzles}} \\
baba\_\allowbreak{}is\_\allowbreak{}ai/\allowbreak{}goto\_\allowbreak{}win & Baba Is AI env/goto\_win (\citealp{cloos2024baba}; baba-is-ai at 40ae1b2), a BALROG task & paused & 3 & 100 & progression (\%: 100 if solved) & 0.00 / 0.0 \\
baba\_\allowbreak{}is\_\allowbreak{}ai/\allowbreak{}goto\_\allowbreak{}win-distr\_\allowbreak{}obj & Baba Is AI env/goto\_win-distr\_obj (\citealp{cloos2024baba}; baba-is-ai at 40ae1b2), a BALROG task & paused & 3 & 100 & progression (\%: 100 if solved) & 0.33 / 33.3 \\
baba\_\allowbreak{}is\_\allowbreak{}ai/\allowbreak{}goto\_\allowbreak{}win-distr\_\allowbreak{}obj-irrelevant\_\allowbreak{}rule & Baba Is AI env/goto\_win-distr\_obj-irrelevant\_rule (\citealp{cloos2024baba}; baba-is-ai at 40ae1b2), a BALROG task & paused & 3 & 100 & progression (\%: 100 if solved) & 0.00 / 0.0 \\
baba\_\allowbreak{}is\_\allowbreak{}ai/\allowbreak{}goto\_\allowbreak{}win-distr\_\allowbreak{}obj\_\allowbreak{}rule & Baba Is AI env/goto\_win-distr\_obj\_rule (\citealp{cloos2024baba}; baba-is-ai at 40ae1b2), a BALROG task & paused & 3 & 100 & progression (\%: 100 if solved) & 0.33 / 33.3 \\
baba\_\allowbreak{}is\_\allowbreak{}ai/\allowbreak{}goto\_\allowbreak{}win-distr\_\allowbreak{}rule & Baba Is AI env/goto\_win-distr\_rule (\citealp{cloos2024baba}; baba-is-ai at 40ae1b2), a BALROG task & paused & 3 & 100 & progression (\%: 100 if solved) & 0.00 / 0.0 \\
baba\_\allowbreak{}is\_\allowbreak{}ai/\allowbreak{}make\_\allowbreak{}win & Baba Is AI env/make\_win (\citealp{cloos2024baba}; baba-is-ai at 40ae1b2), a BALROG task & paused & 3 & 100 & progression (\%: 100 if solved) & 0.00 / 0.0 \\
baba\_\allowbreak{}is\_\allowbreak{}ai/\allowbreak{}make\_\allowbreak{}win-distr\_\allowbreak{}obj & Baba Is AI env/make\_win-distr\_obj (\citealp{cloos2024baba}; baba-is-ai at 40ae1b2), a BALROG task & paused & 3 & 100 & progression (\%: 100 if solved) & 0.00 / 0.0 \\
baba\_\allowbreak{}is\_\allowbreak{}ai/\allowbreak{}make\_\allowbreak{}win-distr\_\allowbreak{}obj-irrelevant\_\allowbreak{}rule & Baba Is AI env/make\_win-distr\_obj-irrelevant\_rule (\citealp{cloos2024baba}; baba-is-ai at 40ae1b2), a BALROG task & paused & 3 & 100 & progression (\%: 100 if solved) & 0.00 / 0.0 \\
baba\_\allowbreak{}is\_\allowbreak{}ai/\allowbreak{}make\_\allowbreak{}win-distr\_\allowbreak{}obj\_\allowbreak{}rule & Baba Is AI env/make\_win-distr\_obj\_rule (\citealp{cloos2024baba}; baba-is-ai at 40ae1b2), a BALROG task & paused & 3 & 100 & progression (\%: 100 if solved) & 0.00 / 0.0 \\
baba\_\allowbreak{}is\_\allowbreak{}ai/\allowbreak{}make\_\allowbreak{}win-distr\_\allowbreak{}rule & Baba Is AI env/make\_win-distr\_rule (\citealp{cloos2024baba}; baba-is-ai at 40ae1b2), a BALROG task & paused & 3 & 100 & progression (\%: 100 if solved) & 0.00 / 0.0 \\
baba\_\allowbreak{}is\_\allowbreak{}ai/\allowbreak{}two\_\allowbreak{}room-break\_\allowbreak{}stop-goto\_\allowbreak{}win & Baba Is AI env/two\_room-break\_stop-goto\_win (\citealp{cloos2024baba}; baba-is-ai at 40ae1b2), a BALROG task & paused & 3 & 100 & progression (\%: 100 if solved) & 0.00 / 0.0 \\
baba\_\allowbreak{}is\_\allowbreak{}ai/\allowbreak{}two\_\allowbreak{}room-break\_\allowbreak{}stop-goto\_\allowbreak{}win-distr\_\allowbreak{}obj & Baba Is AI env/two\_room-break\_stop-goto\_win-distr\_obj (\citealp{cloos2024baba}; baba-is-ai at 40ae1b2), a BALROG task & paused & 3 & 100 & progression (\%: 100 if solved) & 0.00 / 0.0 \\
baba\_\allowbreak{}is\_\allowbreak{}ai/\allowbreak{}two\_\allowbreak{}room-break\_\allowbreak{}stop-goto\_\allowbreak{}win-distr\_\allowbreak{}obj-irrelevant\_\allowbreak{}rule & Baba Is AI env/two\_room-break\_stop-goto\_win-distr\_obj-irrelevant\_rule (\citealp{cloos2024baba}; baba-is-ai at 40ae1b2), a BALROG task & paused & 3 & 100 & progression (\%: 100 if solved) & 0.00 / 0.0 \\
baba\_\allowbreak{}is\_\allowbreak{}ai/\allowbreak{}two\_\allowbreak{}room-break\_\allowbreak{}stop-goto\_\allowbreak{}win-distr\_\allowbreak{}obj\_\allowbreak{}rule & Baba Is AI env/two\_room-break\_stop-goto\_win-distr\_obj\_rule (\citealp{cloos2024baba}; baba-is-ai at 40ae1b2), a BALROG task & paused & 3 & 100 & progression (\%: 100 if solved) & 0.00 / 0.0 \\
baba\_\allowbreak{}is\_\allowbreak{}ai/\allowbreak{}two\_\allowbreak{}room-break\_\allowbreak{}stop-goto\_\allowbreak{}win-distr\_\allowbreak{}rule & Baba Is AI env/two\_room-break\_stop-goto\_win-distr\_rule (\citealp{cloos2024baba}; baba-is-ai at 40ae1b2), a BALROG task & paused & 3 & 100 & progression (\%: 100 if solved) & 0.00 / 0.0 \\
baba\_\allowbreak{}is\_\allowbreak{}ai/\allowbreak{}two\_\allowbreak{}room-break\_\allowbreak{}stop-make\_\allowbreak{}win & Baba Is AI env/two\_room-break\_stop-make\_win (\citealp{cloos2024baba}; baba-is-ai at 40ae1b2), a BALROG task & paused & 3 & 100 & progression (\%: 100 if solved) & 0.00 / 0.0 \\
baba\_\allowbreak{}is\_\allowbreak{}ai/\allowbreak{}two\_\allowbreak{}room-break\_\allowbreak{}stop-make\_\allowbreak{}win-distr\_\allowbreak{}obj & Baba Is AI env/two\_room-break\_stop-make\_win-distr\_obj (\citealp{cloos2024baba}; baba-is-ai at 40ae1b2), a BALROG task & paused & 3 & 100 & progression (\%: 100 if solved) & 0.00 / 0.0 \\
baba\_\allowbreak{}is\_\allowbreak{}ai/\allowbreak{}two\_\allowbreak{}room-break\_\allowbreak{}stop-make\_\allowbreak{}win-distr\_\allowbreak{}obj-irrelevant\_\allowbreak{}rule & Baba Is AI env/two\_room-break\_stop-make\_win-distr\_obj-irrelevant\_rule (\citealp{cloos2024baba}; baba-is-ai at 40ae1b2), a BALROG task & paused & 3 & 100 & progression (\%: 100 if solved) & 0.00 / 0.0 \\
baba\_\allowbreak{}is\_\allowbreak{}ai/\allowbreak{}two\_\allowbreak{}room-break\_\allowbreak{}stop-make\_\allowbreak{}win-distr\_\allowbreak{}obj\_\allowbreak{}rule & Baba Is AI env/two\_room-break\_stop-make\_win-distr\_obj\_rule (\citealp{cloos2024baba}; baba-is-ai at 40ae1b2), a BALROG task & paused & 3 & 100 & progression (\%: 100 if solved) & 0.00 / 0.0 \\
baba\_\allowbreak{}is\_\allowbreak{}ai/\allowbreak{}two\_\allowbreak{}room-break\_\allowbreak{}stop-make\_\allowbreak{}win-distr\_\allowbreak{}rule & Baba Is AI env/two\_room-break\_stop-make\_win-distr\_rule (\citealp{cloos2024baba}; baba-is-ai at 40ae1b2), a BALROG task & paused & 3 & 100 & progression (\%: 100 if solved) & 0.00 / 0.0 \\
baba\_\allowbreak{}is\_\allowbreak{}ai/\allowbreak{}two\_\allowbreak{}room-goto\_\allowbreak{}win & Baba Is AI env/two\_room-goto\_win (\citealp{cloos2024baba}; baba-is-ai at 40ae1b2), a BALROG task & paused & 3 & 100 & progression (\%: 100 if solved) & 0.33 / 33.3 \\
baba\_\allowbreak{}is\_\allowbreak{}ai/\allowbreak{}two\_\allowbreak{}room-goto\_\allowbreak{}win-distr\_\allowbreak{}obj & Baba Is AI env/two\_room-goto\_win-distr\_obj (\citealp{cloos2024baba}; baba-is-ai at 40ae1b2), a BALROG task & paused & 3 & 100 & progression (\%: 100 if solved) & 0.33 / 33.3 \\
baba\_\allowbreak{}is\_\allowbreak{}ai/\allowbreak{}two\_\allowbreak{}room-goto\_\allowbreak{}win-distr\_\allowbreak{}obj-irrelevant\_\allowbreak{}rule & Baba Is AI env/two\_room-goto\_win-distr\_obj-irrelevant\_rule (\citealp{cloos2024baba}; baba-is-ai at 40ae1b2), a BALROG task & paused & 3 & 100 & progression (\%: 100 if solved) & 0.67 / 66.7 \\
baba\_\allowbreak{}is\_\allowbreak{}ai/\allowbreak{}two\_\allowbreak{}room-goto\_\allowbreak{}win-distr\_\allowbreak{}obj\_\allowbreak{}rule & Baba Is AI env/two\_room-goto\_win-distr\_obj\_rule (\citealp{cloos2024baba}; baba-is-ai at 40ae1b2), a BALROG task & paused & 3 & 100 & progression (\%: 100 if solved) & 0.33 / 33.3 \\
\bottomrule
\end{tabular}
\end{table*}

\begin{table*}[p]
\centering
\scriptsize
\setlength{\tabcolsep}{3pt}
\caption*{Table~\ref{tab:envs} (continued).}
\begin{tabular}{@{}>{\raggedright\arraybackslash}p{3.4cm}>{\raggedright\arraybackslash}p{5.4cm}lrr>{\raggedright\arraybackslash}p{1.5cm}r@{}}
\toprule
Environment & Source & Mode & Episodes & Max dec. & Score & Jev \\
\midrule
baba\_\allowbreak{}is\_\allowbreak{}ai/\allowbreak{}two\_\allowbreak{}room-goto\_\allowbreak{}win-distr\_\allowbreak{}rule & Baba Is AI env/two\_room-goto\_win-distr\_rule (\citealp{cloos2024baba}; baba-is-ai at 40ae1b2), a BALROG task & paused & 3 & 100 & progression (\%: 100 if solved) & 0.67 / 66.7 \\
baba\_\allowbreak{}is\_\allowbreak{}ai/\allowbreak{}two\_\allowbreak{}room-goto\_\allowbreak{}win-distr\_\allowbreak{}win\_\allowbreak{}rule & Baba Is AI env/two\_room-goto\_win-distr\_win\_rule (\citealp{cloos2024baba}; baba-is-ai at 40ae1b2), a BALROG task & paused & 3 & 100 & progression (\%: 100 if solved) & 0.33 / 33.3 \\
baba\_\allowbreak{}is\_\allowbreak{}ai/\allowbreak{}two\_\allowbreak{}room-make\_\allowbreak{}wall\_\allowbreak{}win & Baba Is AI env/two\_room-make\_wall\_win (\citealp{cloos2024baba}; baba-is-ai at 40ae1b2), a BALROG task & paused & 3 & 100 & progression (\%: 100 if solved) & 0.00 / 0.0 \\
baba\_\allowbreak{}is\_\allowbreak{}ai/\allowbreak{}two\_\allowbreak{}room-make\_\allowbreak{}win & Baba Is AI env/two\_room-make\_win (\citealp{cloos2024baba}; baba-is-ai at 40ae1b2), a BALROG task & paused & 3 & 100 & progression (\%: 100 if solved) & 0.00 / 0.0 \\
baba\_\allowbreak{}is\_\allowbreak{}ai/\allowbreak{}two\_\allowbreak{}room-make\_\allowbreak{}win-distr\_\allowbreak{}obj & Baba Is AI env/two\_room-make\_win-distr\_obj (\citealp{cloos2024baba}; baba-is-ai at 40ae1b2), a BALROG task & paused & 3 & 100 & progression (\%: 100 if solved) & 0.00 / 0.0 \\
baba\_\allowbreak{}is\_\allowbreak{}ai/\allowbreak{}two\_\allowbreak{}room-make\_\allowbreak{}win-distr\_\allowbreak{}obj-irrelevant\_\allowbreak{}rule & Baba Is AI env/two\_room-make\_win-distr\_obj-irrelevant\_rule (\citealp{cloos2024baba}; baba-is-ai at 40ae1b2), a BALROG task & paused & 3 & 100 & progression (\%: 100 if solved) & 0.00 / 0.0 \\
baba\_\allowbreak{}is\_\allowbreak{}ai/\allowbreak{}two\_\allowbreak{}room-make\_\allowbreak{}win-distr\_\allowbreak{}obj\_\allowbreak{}rule & Baba Is AI env/two\_room-make\_win-distr\_obj\_rule (\citealp{cloos2024baba}; baba-is-ai at 40ae1b2), a BALROG task & paused & 3 & 100 & progression (\%: 100 if solved) & 0.00 / 0.0 \\
baba\_\allowbreak{}is\_\allowbreak{}ai/\allowbreak{}two\_\allowbreak{}room-make\_\allowbreak{}win-distr\_\allowbreak{}rule & Baba Is AI env/two\_room-make\_win-distr\_rule (\citealp{cloos2024baba}; baba-is-ai at 40ae1b2), a BALROG task & paused & 3 & 100 & progression (\%: 100 if solved) & 0.00 / 0.0 \\
baba\_\allowbreak{}is\_\allowbreak{}ai/\allowbreak{}two\_\allowbreak{}room-make\_\allowbreak{}win-distr\_\allowbreak{}win\_\allowbreak{}rule & Baba Is AI env/two\_room-make\_win-distr\_win\_rule (\citealp{cloos2024baba}; baba-is-ai at 40ae1b2), a BALROG task & paused & 3 & 100 & progression (\%: 100 if solved) & 0.00 / 0.0 \\
baba\_\allowbreak{}is\_\allowbreak{}ai/\allowbreak{}two\_\allowbreak{}room-make\_\allowbreak{}you & Baba Is AI env/two\_room-make\_you (\citealp{cloos2024baba}; baba-is-ai at 40ae1b2), a BALROG task & paused & 3 & 100 & progression (\%: 100 if solved) & 0.00 / 0.0 \\
baba\_\allowbreak{}is\_\allowbreak{}ai/\allowbreak{}two\_\allowbreak{}room-make\_\allowbreak{}you-make\_\allowbreak{}win & Baba Is AI env/two\_room-make\_you-make\_win (\citealp{cloos2024baba}; baba-is-ai at 40ae1b2), a BALROG task & paused & 3 & 100 & progression (\%: 100 if solved) & 0.00 / 0.0 \\
baba\_\allowbreak{}is\_\allowbreak{}ai/\allowbreak{}two\_\allowbreak{}room-maybe\_\allowbreak{}break\_\allowbreak{}stop-goto\_\allowbreak{}win & Baba Is AI env/two\_room-maybe\_break\_stop-goto\_win (\citealp{cloos2024baba}; baba-is-ai at 40ae1b2), a BALROG task & paused & 3 & 100 & progression (\%: 100 if solved) & 0.00 / 0.0 \\
baba\_\allowbreak{}is\_\allowbreak{}ai/\allowbreak{}two\_\allowbreak{}room-maybe\_\allowbreak{}break\_\allowbreak{}stop-goto\_\allowbreak{}win-distr\_\allowbreak{}obj & Baba Is AI env/two\_room-maybe\_break\_stop-goto\_win-distr\_obj (\citealp{cloos2024baba}; baba-is-ai at 40ae1b2), a BALROG task & paused & 3 & 100 & progression (\%: 100 if solved) & 0.00 / 0.0 \\
baba\_\allowbreak{}is\_\allowbreak{}ai/\allowbreak{}two\_\allowbreak{}room-maybe\_\allowbreak{}break\_\allowbreak{}stop-goto\_\allowbreak{}win-distr\_\allowbreak{}obj-irrelevant\_\allowbreak{}rule & Baba Is AI env/two\_room-maybe\_break\_stop-goto\_win-distr\_obj-irrelevant\_rule (\citealp{cloos2024baba}; baba-is-ai at 40ae1b2), a BALROG task & paused & 3 & 100 & progression (\%: 100 if solved) & 0.33 / 33.3 \\
baba\_\allowbreak{}is\_\allowbreak{}ai/\allowbreak{}two\_\allowbreak{}room-maybe\_\allowbreak{}break\_\allowbreak{}stop-goto\_\allowbreak{}win-distr\_\allowbreak{}obj\_\allowbreak{}rule & Baba Is AI env/two\_room-maybe\_break\_stop-goto\_win-distr\_obj\_rule (\citealp{cloos2024baba}; baba-is-ai at 40ae1b2), a BALROG task & paused & 3 & 100 & progression (\%: 100 if solved) & 0.00 / 0.0 \\
baba\_\allowbreak{}is\_\allowbreak{}ai/\allowbreak{}two\_\allowbreak{}room-maybe\_\allowbreak{}break\_\allowbreak{}stop-goto\_\allowbreak{}win-distr\_\allowbreak{}rule & Baba Is AI env/two\_room-maybe\_break\_stop-goto\_win-distr\_rule (\citealp{cloos2024baba}; baba-is-ai at 40ae1b2), a BALROG task & paused & 3 & 100 & progression (\%: 100 if solved) & 0.33 / 33.3 \\
\multicolumn{7}{@{}l}{\textbf{loop/coding\_controller}} \\
swebench\_\allowbreak{}verified\_\allowbreak{}gpt41mini & SWE-bench Verified (\citealp{jimenez2024swebench}; \citealp{openai2024sweverified}), official evaluation images and swebench 5.0.2 grading & paused & 40 & 30 & resolved (1) or not (0) & 0.12 / 0.1 \\
swebench\_\allowbreak{}verified\_\allowbreak{}gpt56luna & SWE-bench Verified (\citealp{jimenez2024swebench}; \citealp{openai2024sweverified}), official evaluation images and swebench 5.0.2 grading & paused & 40 & 30 & resolved (1) or not (0) & -- \\
swebench\_\allowbreak{}verified\_\allowbreak{}scripted & SWE-bench Verified (\citealp{jimenez2024swebench}; \citealp{openai2024sweverified}), official evaluation images and swebench 5.0.2 grading & paused & 40 & 30 & resolved (1) or not (0) & 0.40 / 0.4 \\
\multicolumn{7}{@{}l}{\textbf{loop/customer\_service}} \\
tau2\_\allowbreak{}telecom\_\allowbreak{}solo & $\tau$$^2$-bench (\citealp{barres2025tau2}), telecom domain, no-user (solo) mode & paused & 114 & 200 & reward & 0.91 / 0.9 \\
\multicolumn{7}{@{}l}{\textbf{loop/science\_experiments}} \\
scienceworld & ScienceWorld (\citealp{wang2022scienceworld}), 30 task types, test variations, 'easy' simplifications & paused & 30 & 300 & task score (0-100) & 0.27 / 38.4 \\
\multicolumn{7}{@{}l}{\textbf{loop/text\_games}} \\
jericho\_\allowbreak{}adventureland & Jericho (\citealp{hausknecht2020jericho}) game suite: Adventureland (Scott Adams, 1978; Inform port, 1996) & paused & 3 & 100 & points & 0.00 / 0.0 \\
jericho\_\allowbreak{}deephome & Jericho (\citealp{hausknecht2020jericho}) game suite: Deephome: A Telleen Adventure (Joshua Wise, 1999) & paused & 3 & 100 & points & 0.00 / 1.0 \\
\bottomrule
\end{tabular}
\end{table*}

\begin{table*}[p]
\centering
\scriptsize
\setlength{\tabcolsep}{3pt}
\caption*{Table~\ref{tab:envs} (continued).}
\begin{tabular}{@{}>{\raggedright\arraybackslash}p{3.4cm}>{\raggedright\arraybackslash}p{5.4cm}lrr>{\raggedright\arraybackslash}p{1.5cm}r@{}}
\toprule
Environment & Source & Mode & Episodes & Max dec. & Score & Jev \\
\midrule
jericho\_\allowbreak{}detective & Jericho (\citealp{hausknecht2020jericho}) game suite: Detective (Matt Barringer, ported by Stuart Moore, 2000) & paused & 1 & 100 & points & 0.00 / 30.0 \\
jericho\_\allowbreak{}dragon & Jericho (\citealp{hausknecht2020jericho}) game suite: Dragon Adventure (William Stott) & paused & 3 & 100 & points & 0.00 / -13.3 \\
jericho\_\allowbreak{}library & Jericho (\citealp{hausknecht2020jericho}) game suite: All Quiet on the Library Front (Michael S. Phillips, 1995) & paused & 1 & 100 & points & 0.00 / 0.0 \\
jericho\_\allowbreak{}ludicorp & Jericho (\citealp{hausknecht2020jericho}) game suite: The Ludicorp Mystery (Cal Henderson, 2003) & paused & 1 & 100 & points & 0.00 / 5.0 \\
jericho\_\allowbreak{}omniquest & Jericho (\citealp{hausknecht2020jericho}) game suite: OMNIQuest (Chris Barden and Chris Ethridge, c. 1988; Inform port) & paused & 1 & 100 & points & 0.00 / 0.0 \\
jericho\_\allowbreak{}pentari & Jericho (\citealp{hausknecht2020jericho}) game suite: Pentari (Howard A. Sherman, 1998) & paused & 3 & 100 & points & 0.00 / 0.0 \\
jericho\_\allowbreak{}zork1 & Jericho (\citealp{hausknecht2020jericho}) game suite: Zork I: The Great Underground Empire (Infocom, 1980) & paused & 3 & 100 & points & 0.00 / 0.0 \\
jericho\_\allowbreak{}zork3 & Jericho (\citealp{hausknecht2020jericho}) game suite: Zork III: The Dungeon Master (Infocom, 1982) & paused & 3 & 100 & points & 0.00 / 1.0 \\
tw\_\allowbreak{}cooking\_\allowbreak{}level3 & TextWorld cooking game (First TextWorld Problems generator), GATA test set, difficulty level 3: 1 ingredient to find and cut & paused & 20 & 50 & points & 0.55 / 3.1 \\
tw\_\allowbreak{}cooking\_\allowbreak{}level5 & TextWorld cooking game (First TextWorld Problems generator), GATA test set, difficulty level 5: 1 ingredient to find & paused & 20 & 100 & points & 0.60 / 1.9 \\
tw\_\allowbreak{}cooking\_\allowbreak{}level7 & TextWorld cooking game (First TextWorld Problems generator), GATA test set, difficulty level 7: 1 ingredient to find, cut and cook & paused & 20 & 50 & points & 0.95 / 4.8 \\
tw\_\allowbreak{}cooking\_\allowbreak{}level9 & TextWorld cooking game (First TextWorld Problems generator), GATA test set, difficulty level 9: 3 ingredients to find, cut and cook & paused & 20 & 100 & points & 0.50 / 8.3 \\
\multicolumn{7}{@{}l}{\textbf{loop/web}} \\
miniwob/\allowbreak{}book-flight-nodelay & MiniWoB++ book-flight-nodelay (\citealp{liu2018miniwob}; miniwob 1.1) & paused & 5 & 25 & reward & 0.00 / -1.0 \\
miniwob/\allowbreak{}buy-ticket & MiniWoB++ buy-ticket (\citealp{liu2018miniwob}; miniwob 1.1) & paused & 5 & 5 & reward & 1.00 / 1.0 \\
miniwob/\allowbreak{}choose-date-nodelay & MiniWoB++ choose-date-nodelay (\citealp{liu2018miniwob}; miniwob 1.1) & paused & 5 & 30 & reward & 0.20 / -0.6 \\
miniwob/\allowbreak{}choose-list & MiniWoB++ choose-list (\citealp{liu2018miniwob}; miniwob 1.1) & paused & 5 & 6 & reward & 1.00 / 1.0 \\
miniwob/\allowbreak{}click-button & MiniWoB++ click-button (\citealp{liu2018miniwob}; miniwob 1.1) & paused & 5 & 5 & reward & 1.00 / 1.0 \\
miniwob/\allowbreak{}click-button-sequence & MiniWoB++ click-button-sequence (\citealp{liu2018miniwob}; miniwob 1.1) & paused & 5 & 6 & reward & 1.00 / 1.0 \\
miniwob/\allowbreak{}click-checkboxes-large & MiniWoB++ click-checkboxes-large (\citealp{liu2018miniwob}; miniwob 1.1) & paused & 5 & 25 & reward & 1.00 / 0.9 \\
miniwob/\allowbreak{}click-checkboxes-soft & MiniWoB++ click-checkboxes-soft (\citealp{liu2018miniwob}; miniwob 1.1) & paused & 5 & 12 & reward & 1.00 / 0.9 \\
miniwob/\allowbreak{}click-collapsible-2-nodelay & MiniWoB++ click-collapsible-2-nodelay (\citealp{liu2018miniwob}; miniwob 1.1) & paused & 5 & 10 & reward & 1.00 / 1.0 \\
miniwob/\allowbreak{}click-dialog-2 & MiniWoB++ click-dialog-2 (\citealp{liu2018miniwob}; miniwob 1.1) & paused & 5 & 5 & reward & 0.80 / 0.6 \\
miniwob/\allowbreak{}click-option & MiniWoB++ click-option (\citealp{liu2018miniwob}; miniwob 1.1) & paused & 5 & 6 & reward & 1.00 / 1.0 \\
miniwob/\allowbreak{}click-scroll-list & MiniWoB++ click-scroll-list (\citealp{liu2018miniwob}; miniwob 1.1) & paused & 5 & 12 & reward & 0.80 / 0.6 \\
miniwob/\allowbreak{}click-tab-2-hard & MiniWoB++ click-tab-2-hard (\citealp{liu2018miniwob}; miniwob 1.1) & paused & 5 & 10 & reward & 1.00 / 1.0 \\
miniwob/\allowbreak{}click-widget & MiniWoB++ click-widget (\citealp{liu2018miniwob}; miniwob 1.1) & paused & 5 & 5 & reward & 1.00 / 1.0 \\
miniwob/\allowbreak{}copy-paste-2 & MiniWoB++ copy-paste-2 (\citealp{liu2018miniwob}; miniwob 1.1) & paused & 5 & 6 & reward & 0.00 / -1.0 \\
miniwob/\allowbreak{}email-inbox & MiniWoB++ email-inbox (\citealp{liu2018miniwob}; miniwob 1.1) & paused & 5 & 10 & reward & 0.20 / -0.6 \\
miniwob/\allowbreak{}email-inbox-forward-nl-turk & MiniWoB++ email-inbox-forward-nl-turk (\citealp{liu2018miniwob}; miniwob 1.1) & paused & 5 & 10 & reward & 0.00 / -1.0 \\
\bottomrule
\end{tabular}
\end{table*}

\begin{table*}[p]
\centering
\scriptsize
\setlength{\tabcolsep}{3pt}
\caption*{Table~\ref{tab:envs} (continued).}
\begin{tabular}{@{}>{\raggedright\arraybackslash}p{3.4cm}>{\raggedright\arraybackslash}p{5.4cm}lrr>{\raggedright\arraybackslash}p{1.5cm}r@{}}
\toprule
Environment & Source & Mode & Episodes & Max dec. & Score & Jev \\
\midrule
miniwob/\allowbreak{}enter-date & MiniWoB++ enter-date (\citealp{liu2018miniwob}; miniwob 1.1) & paused & 5 & 6 & reward & 0.00 / -1.0 \\
miniwob/\allowbreak{}enter-password & MiniWoB++ enter-password (\citealp{liu2018miniwob}; miniwob 1.1) & paused & 5 & 8 & reward & 0.00 / -1.0 \\
miniwob/\allowbreak{}form-sequence-2 & MiniWoB++ form-sequence-2 (\citealp{liu2018miniwob}; miniwob 1.1) & paused & 5 & 8 & reward & 0.00 / -1.0 \\
miniwob/\allowbreak{}login-user-popup & MiniWoB++ login-user-popup (\citealp{liu2018miniwob}; miniwob 1.1) & paused & 5 & 12 & reward & 0.00 / -1.0 \\
miniwob/\allowbreak{}multi-layouts & MiniWoB++ multi-layouts (\citealp{liu2018miniwob}; miniwob 1.1) & paused & 5 & 10 & reward & 0.00 / -1.0 \\
miniwob/\allowbreak{}navigate-tree & MiniWoB++ navigate-tree (\citealp{liu2018miniwob}; miniwob 1.1) & paused & 5 & 10 & reward & 1.00 / 1.0 \\
miniwob/\allowbreak{}phone-book & MiniWoB++ phone-book (\citealp{liu2018miniwob}; miniwob 1.1) & paused & 5 & 15 & reward & 0.80 / 0.7 \\
miniwob/\allowbreak{}read-table-2 & MiniWoB++ read-table-2 (\citealp{liu2018miniwob}; miniwob 1.1) & paused & 5 & 8 & reward & 0.00 / -1.0 \\
miniwob/\allowbreak{}search-engine & MiniWoB++ search-engine (\citealp{liu2018miniwob}; miniwob 1.1) & paused & 5 & 10 & reward & 0.20 / -0.6 \\
miniwob/\allowbreak{}social-media & MiniWoB++ social-media (\citealp{liu2018miniwob}; miniwob 1.1) & paused & 5 & 6 & reward & 1.00 / 1.0 \\
miniwob/\allowbreak{}social-media-some & MiniWoB++ social-media-some (\citealp{liu2018miniwob}; miniwob 1.1) & paused & 5 & 20 & reward & 0.80 / 0.6 \\
miniwob/\allowbreak{}tic-tac-toe & MiniWoB++ tic-tac-toe (\citealp{liu2018miniwob}; miniwob 1.1) & paused & 5 & 10 & reward & 0.40 / 0.1 \\
miniwob/\allowbreak{}use-autocomplete-nodelay & MiniWoB++ use-autocomplete-nodelay (\citealp{liu2018miniwob}; miniwob 1.1) & paused & 5 & 8 & reward & 0.00 / -1.0 \\
\multicolumn{7}{@{}l}{\textbf{loop/web\_shopping}} \\
webshop & WebShop (\citealp{yao2022webshop}), text mode, all 1.18M products, human instructions (test split) & paused & 100 & 15 & reward & 0.29 / 0.6 \\
\multicolumn{7}{@{}l}{\textbf{loop/wikiracing}} \\
wikispeedia & Wikispeedia (\citealp{west2012human}): 2007 Wikipedia Selection for Schools, its link graph and finished human games & paused & 20 & 30 & clicks (lower is better) & 0.85 / 6.6 \\
\multicolumn{7}{@{}l}{\textbf{realtime/atari}} \\
boxing & Boxing (Atari 2600) in the Arcade Learning Environment, objects from OCAtari & realtime & 10 & 1787 & game score & -- / -16.1 \\
breakout & Breakout (Atari 2600) in the Arcade Learning Environment, objects from OCAtari & realtime & 10 & 1800 & game score & -- / 1.9 \\
freeway & Freeway (Atari 2600) in the Arcade Learning Environment, objects from OCAtari & realtime & 10 & 2049 & game score & -- / 20.0 \\
pong & Pong (Atari 2600) in the Arcade Learning Environment, objects from OCAtari & realtime & 10 & 1800 & game score & -- / -21.0 \\
\multicolumn{7}{@{}l}{\textbf{realtime/doom}} \\
basic & ViZDoom basic scenario (\citealp{kempka2016vizdoom}) & realtime & 10 & 75 & return & 0.30 / -285.2 \\
deadly\_\allowbreak{}corridor & ViZDoom deadly\_corridor scenario (\citealp{kempka2016vizdoom}) & realtime & 10 & 525 & return & 0.00 / 183.0 \\
defend\_\allowbreak{}the\_\allowbreak{}center & ViZDoom defend\_the\_center scenario (\citealp{kempka2016vizdoom}) & realtime & 10 & 525 & return & -- / 1.5 \\
health\_\allowbreak{}gathering & ViZDoom health\_gathering scenario (\citealp{kempka2016vizdoom}) & realtime & 10 & 525 & return & 0.00 / 418.4 \\
\multicolumn{7}{@{}l}{\textbf{realtime/drone}} \\
quadx\_\allowbreak{}waypoints & PyFlyt QuadX-Waypoints-v4 (\citealp{tai2023pyflyt}) & realtime & 10 & 902 & return & 0.00 / -70.2 \\
\multicolumn{7}{@{}l}{\textbf{realtime/highway}} \\
highway & highway-env highway-v0 (\citealp{leurent2018highway}) & realtime & 10 & 40 & return & 0.00 / 7.1 \\
intersection & highway-env intersection-v0 (\citealp{leurent2018highway}) & realtime & 10 & 13 & return & 0.30 / 0.6 \\
intersection\_\allowbreak{}multi\_\allowbreak{}agent & highway-env intersection-multi-agent-v0 (\citealp{leurent2018highway}), destinations changed so that the paths conflict & realtime & 10 & 20 & return & 0.10 / 3.9 \\
merge & highway-env merge-v0 (\citealp{leurent2018highway}) & realtime & 10 & 30 & return & 0.00 / 4.9 \\
roundabout & highway-env roundabout-v0 (\citealp{leurent2018highway}) & realtime & 10 & 11 & return & 0.70 / 8.5 \\
\bottomrule
\end{tabular}
\end{table*}

\end{document}